\documentclass[a4paper,fleqn]{cas-sc}

\usepackage{graphicx}
\usepackage{float}
\usepackage{capt-of}
\usepackage{url}
\usepackage{booktabs}
\usepackage{tabularx}
\usepackage{multirow}
\usepackage[table,xcdraw]{xcolor}
\usepackage{amsmath,amssymb,amsfonts}
\usepackage{algorithm}
\usepackage{algpseudocode}
\usepackage{orcidlink}
\usepackage{xcolor}
\usepackage[numbers,sort&compress]{natbib}

\def\tsc#1{\csdef{#1}{\textsc{\lowercase{#1}}\xspace}}
\tsc{WGM}
\tsc{QE}

\begin{document}
\let\WriteBookmarks\relax
\let\printorcid\relax
\def\floatpagepagefraction{1}
\def\textpagefraction{.001}

\shorttitle{AVA-VLM: Adaptive Visual Attention-Vision Language Model for In-the-Wild Construction Site Monitoring}    

\shortauthors{Y. Kim et al.} 

\title [mode = title]{AVA-VLM: Adaptive Visual Attention-Vision Language Model for In-the-Wild Construction Site Monitoring}

\author[1, 2]{Younggun Kim\,\orcidlink{0009-0005-3108-5304}}
\author[3]{Taeheon Kim}
\author[1]{Youngseo Kim}
\author[2, 3, 4]{Seunghee Park\,\orcidlink{0000-0002-6265-7652}}
\cormark[1]

\fntext[1]{%
E-mail addresses: 
kyg9191@ucla.edu (Y. Kim); 
taeheon@smartinside.ai (T. Kim);
youngseo@ucla.edu (Y. Kim); 
shparkpc@skku.edu (S. Park).
}

\affiliation[1]{organization={Department of Civil and Environmental Engineering, University of California, Los Angeles}, 
            city={405 Hilgard Avenue, Los Angeles},
            postcode={90095}, 
            state={CA},
            country={USA}}

\affiliation[2]{organization={AI+KCIR Global Resilience Research Center},
            city={Suwon},
            postcode={16419},
            state={Gyeonggi-do},
            country={South Korea}}

\affiliation[3]{organization={SmartInside AI Co., Ltd.},
            city={Suwon},
            postcode={16419}, 
            state={Gyeonggi-do},
            country={South Korea}}

\affiliation[4]{organization={School of Civil, Architectural Engineering and Landscape Architecture, Sungkyunkwan University},
            city={Suwon},
            postcode={16419},
            state={Gyeonggi-do},
            country={South Korea}}

\cortext[1]{Corresponding author}


\begin{abstract}
Existing construction-site Vision-Language Model (VLM) studies have primarily adapted pretrained VLMs through direct QA-style fine-tuning from a single global image, but we argue that this paradigm remains limited in operational range, reliability under reduced-resolution inputs, and inference efficiency. To address these limitations, we propose AVA-VLM, an Adaptive Visual Attention-Vision Language Model that follows a human-inspired coarse-to-fine strategy: it first reasons over a low-resolution global image and requests a high-resolution local crop only when detailed inspection is needed. We further introduce a region-aware Chain-of-Thought dataset that teaches when to inspect, where to crop, and how to use local evidence. Experiments show that, for violation identification, AVA-VLM improves overall F1 from 62.0 to 75.1 while using only 30.6\% of the baseline visual-token budget; for long-distance PPE-violation cases, F1 improves from 16.0 to 63.6. These results demonstrate AVA-VLM's improved robustness to distant and reduced-resolution visual evidence with substantially lower visual-token usage.
\end{abstract}




\maketitle

\section{Introduction}\label{sec:intro}

Vision-Language Models (VLMs) have emerged as a powerful framework for general-purpose visual understanding. By jointly processing images and natural-language instructions, VLMs can perform diverse tasks such as image captioning, visual question answering (VQA), visual reasoning, and grounded scene interpretation. This flexibility has led to their broad adoption across real-world domains, such as virtual assistants~\cite{virtualassistants1, virtualassistants2}, privacy risk assessment~\cite{privacy1, privacy2}, and transportation safety analysis~\cite{transportation1, transportation2, transportation3}. Recently, construction site monitoring has also emerged as a promising application domain for VLMs, as it requires models to interpret complex visual scenes together with textual safety rules. Unlike conventional vision systems that are typically limited to predefined object categories or narrow detection tasks, VLMs can support a broader range of construction-safety functions, including hazard identification, rule-violation reasoning, visual question answering, visual grounding, and natural-language interaction with site managers.

To build construction-tailored VLMs, recent studies have largely followed a similar development pipeline~\cite{ConstructionContext-aware-vlm, ConstructionCS-VLM, ConstructionDoubleThinking, ConstructionEnhancingVLM_VGR_RL, ConstructionTran1, ConstructionVLMplusYOLO}. First, they curate domain-specific datasets that reflect construction-related tasks, such as personal protective equipment (PPE) compliance, hazard recognition, rule-violation reasoning, or visual question answering. Then, they adapt pretrained VLMs to these datasets through supervised fine-tuning (SFT), often using parameter-efficient fine-tuning (PEFT) methods such as Low-Rank Adaptation (LoRA)~\cite{LoRA} to reduce computational and data requirements while still achieving strong task-specific performance. In most cases, this adaptation follows a direct QA-style paradigm, in which the model receives a single global image and is trained to produce the target response. This line of work has demonstrated the promise of domain adaptation and has shown that construction-specific VLMs can achieve encouraging results on their benchmarks.

Despite these advances, however, we argue that the current adaptation paradigm of construction-tailored VLMs still faces several underexplored challenges for in-the-wild construction-site monitoring, particularly in terms of \textbf{(L1)} operational range, \textbf{(L2)} reliable performance under reduced-resolution inputs, and \textbf{(L3)} inference efficiency, as summarized in Figure~\ref{fig:teaser} (Top).

\begin{center}
\includegraphics[
    width=\textwidth,
    height=\textheight,
    keepaspectratio
]{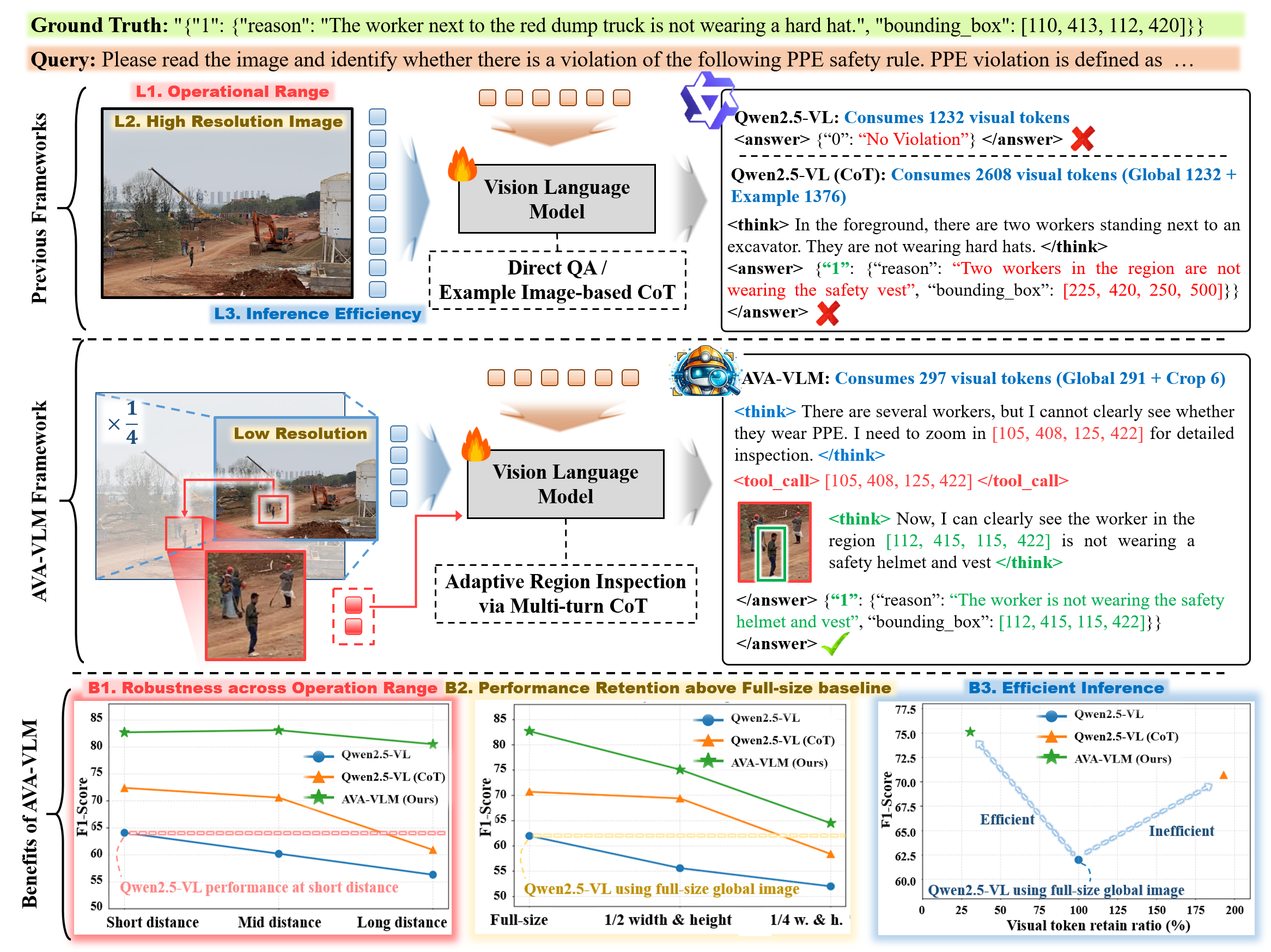}

\captionof{figure}{
Overview of the limitations of existing construction-tailored VLMs and the proposed AVA-VLM.
\textbf{Top:} Existing construction-tailored VLMs rely on full-size visual inputs and direct QA or example image-based Chain-of-Thought (CoT) reasoning, which can limit their operational range (L1), suffer from performance degradation when only reduced-resolution inputs are available during deployment (L2), and increase visual-token consumption (L3).
\textbf{Middle:} AVA-VLM adopts a human-inspired coarse-to-fine reasoning strategy. It first reasons over a low-resolution global image to capture the overall scene context and then adaptively zooms in on a query-relevant region only when detailed inspection is needed.
\textbf{Bottom:} This mechanism provides three key benefits: improved operational range for small or distant objects (B1), reliable performance even when using reduced-resolution global images (B2), and efficient inference by reducing unnecessary global visual-token processing while preserving query-relevant local details (B3).
}
\label{fig:teaser}
\end{center}

First, existing construction-tailored VLMs remain limited in operational range \textbf{(L1)}. General-purpose VLM families such as Qwen~\cite{QwenVL, Qwen2VL, Qwen2_5}, InternVL~\cite{InternVL2_5, InternVL3, InternVL3_5}, and LLaVA~\cite{LLaVA, LLaVA-NeXT, LLaVA-CoT} have demonstrated strong visual understanding capabilities, particularly when target objects are sufficiently large or clearly visible in the input image. However, their performance often degrades substantially when the same objects appear small or distant within wide-view images~\cite{VisualCoT}. This issue is particularly problematic in construction environments, where supervisors often need to monitor wide outdoor areas from far-field viewpoints. For example, in PPE-compliance monitoring, a distant worker may be too small to determine whether the required equipment is present. In such cases, the appropriate action is not to make an immediate judgment, but to recognize that the visual evidence is insufficient and inspect the worker more closely. However, a direct QA-style VLM is still required to produce a definitive answer from the global image, even when the available visual evidence is insufficient for a reliable decision. Despite this limitation, most existing construction-tailored VLMs~\cite{ConstructionContext-aware-vlm, ConstructionCS-VLM, ConstructionDoubleThinking, ConstructionEnhancingVLM_VGR_RL} are adapted to answer directly from a single global image and have been evaluated in controlled, close-range, or narrowly scoped settings. Their limited robustness to increasing target distance is further demonstrated by our distance-stratified evaluations in Tables~\ref{tab:VI_class_distance}, \ref{tab:VI_IoU_distance}, and \ref{tab:OD_distance} and Figure~\ref{fig:VQA_main_example}. Therefore, a construction-site VLM that can explicitly determine when the available visual evidence is insufficient and when and where to zoom in is needed to extend operational range.

Second, the limitation of direct QA-style adaptation becomes more pronounced when the input resolution is reduced \textbf{(L2)}. Open-source VLMs are known to be sensitive to changes in input resolution~\cite{Resolution1, Resolution2}, and this sensitivity is particularly consequential for in-the-wild construction-site monitoring. In practical deployments, camera-feed resolution may vary across environments, and high-resolution images may need to be downsampled because of computational constraints, particularly when VLMs are deployed on edge-computing devices~\cite{Edge1, Edge2}. Under such conditions, query-relevant evidence that already occupies only a small portion of the global image becomes even less visually distinguishable. Because direct QA-style models must still infer the final answer from the reduced-resolution global image, they have no explicit mechanism to recover the local visual details lost during downsampling. As a result, query-relevant visual evidence can become difficult to distinguish, further amplifying the operational-range limitation described in \textbf{L1}. This degradation is reflected in our resolution-stratified evaluation results in Tables~\ref{tab:VI_class_scale}, \ref{tab:VI_IoU_scale}, and \ref{tab:OD_scale} and Figure~\ref{fig:VQA_downsampling_example}. Therefore, a construction-site VLM should not only perform well with high-resolution global images, but also retain access to task-relevant local details when the global input resolution is reduced.


Third, preserving visual details through full-size image processing creates an inference-efficiency bottleneck \textbf{(L3)}. One straightforward way to mitigate the resolution-related degradation in \textbf{L2} is to retain high-resolution inputs so that small or distant evidence remains visually accessible. Accordingly, current construction-tailored VLMs typically process the entire image using dense visual-token representations~\cite{ConstructionContext-aware-vlm, ConstructionCS-VLM, ConstructionDoubleThinking, ConstructionEnhancingVLM_VGR_RL, ConstructionTran1, ConstructionVLMplusYOLO}. However, VLM inference complexity is strongly affected by input sequence length, and visual tokens often constitute a major source of computational overhead~\cite{visualtoken1, VisionThink, AdaptVision}. Processing the full image at high resolution is therefore inherently inefficient when only a small region is relevant to the query. In construction-site scenes, task-relevant evidence may occupy only a small portion of the image, while large regions (e.g., sky, ground, background structures, or empty areas) contribute little to the final prediction. Moreover, when the relevant evidence is already sufficiently visible, a low-resolution global image may be adequate for answering the query, making high-resolution image processing unnecessary. Our overall performance comparisons in Tables~\ref{tab:VI_class_overall}, \ref{tab:VI_IoU_overall}, and \ref{tab:constructionsite_object_detection}, together with Figures~\ref{fig:VQA_main_example} and \ref{fig:VQA_downsampling_example}, show that competitive or improved task performance can be achieved with substantially fewer visual tokens. This indicates that processing the entire image at full resolution for every query is often unnecessary. Therefore, addressing the limitations of direct QA-style construction VLMs requires a mechanism that preserves global scene understanding while allocating high-resolution visual processing only when and where additional local detail is needed.

To collectively address these limitations, we propose Adaptive Visual Attention-Vision Language Model (AVA-VLM), a construction-tailored VLM that follows a human-like visual attention mechanism that progressively shifts from global scene understanding to selective zoom-in inspection. As illustrated in Figure~\ref{fig:teaser} (Middle), AVA-VLM first reasons over a low-resolution global image to capture the overall scene context. When the query-relevant evidence is sufficiently visible, the model directly answers from the global image without zoom-in processing. When detailed inspection is required, the model automatically crops the query-relevant region and incorporates the cropped image into the Chain-of-Thought (CoT) reasoning process, similar to how a human inspector zooms in on hard-to-see yet important areas. This coarse-to-fine design provides three key benefits as shown in Figure~\ref{fig:teaser} (Bottom). First, it extends the operational range of construction-site VLMs by enabling fine-grained reasoning over small or distant objects that are easily overlooked in full-scene images \textbf{(B1)}. Second, it enables reliable performance under reduced-resolution global inputs by combining low-resolution global reasoning with selective high-resolution local inspection, allowing the model to recover query-relevant visual details only when needed \textbf{(B2)}. Third, it reduces the visual token consumption by avoiding unnecessary high-resolution processing for regions that are irrelevant to the query or already sufficiently visible from the global view \textbf{(B3)}.

Our main contributions are summarized as follows:
\begin{itemize}
    \item We propose AVA-VLM, a construction-tailored VLM inspired by human visual attention mechanism, which first scans a low-resolution global image and then adaptively zooms in on query-relevant regions for detailed inspection when needed. Through this, AVA-VLM jointly addresses limitations of construction-tailored VLMs in terms of operational range, reduced-resolution deployment, and inference efficiency.

    \item We introduce a region-aware CoT dataset by extending ConstructionSite10K~\cite{ConstructionSite10K} with adaptive crop annotations and multi-step reasoning traces. The proposed dataset enables the model to learn when detailed inspection is necessary, where to crop, and how to incorporate the cropped visual evidence into the reasoning process.

    \item We provide experimental analysis showing that existing construction-site VLM adaptation strategies exhibit distinct limitations: direct QA-based adaptation is highly sensitive to operational range and input-resolution changes, whereas example image-based CoT improves performance stability but incurs substantial visual-token overhead. In contrast, AVA-VLM achieves reliable performance with fewer visual tokens, improving the overall violation-identification F1 score from 62.0 to 75.1 while using only 30.6\% of the baseline visual-token budget, and increasing long-distance PPE-violation F1 score from 16.0 to 63.6.
\end{itemize}

\section{Literature Review}

\subsection{Adaptation of VLMs for Construction Site Monitoring}

\subsubsection{Dataset-Centric Fine-Tuning for Construction-Site VLMs}

Recent studies on construction-site VLMs have primarily followed a dataset-centric adaptation paradigm~\cite{ConstructionVLM1, ConstructionVLM2, ConstructionVCSQ, ConstructionContext-aware-vlm, ConstructionCS-VLM, ConstructionDoubleThinking, ConstructionEnhancingVLM_VGR_RL, ConstructionTran1, ConstructionVLMplusYOLO}. They first construct domain-specific datasets that reflect construction-site monitoring tasks, such as personal protective equipment (PPE) compliance, hazard recognition, and rule-violation detection. Then, pretrained general-purpose VLMs are adapted to these datasets through supervised fine-tuning (SFT). To reduce the computational burden of full-parameter fine-tuning, many studies further adopt parameter-efficient fine-tuning methods, such as Low-Rank Adaptation (LoRA), which update only a small set of trainable parameters while keeping most of the pretrained model frozen~\cite{ConstructionVLM1, ConstructionVCSQ, ConstructionContext-aware-vlm, ConstructionCS-VLM, ConstructionDoubleThinking}. This line of work has demonstrated the effectiveness of domain adaptation for construction-site understanding. By exposing pretrained VLMs to construction-specific images and textual annotations, fine-tuned models can learn domain-specific objects, safety rules, and response formats more effectively than general-purpose VLMs. In particular, LoRA-based adaptation provides a practical way to specialize large-scale VLMs under limited resources, making it suitable for construction datasets that are often smaller than large-scale general purpose datasets. 

However, a key limitation of this adaptation paradigm is that most existing construction-site VLM studies rely on direct QA supervision~\cite{ConstructionVLM1, ConstructionVLM2, ConstructionVCSQ, ConstructionCS-VLM, ConstructionTran1, ConstructionVLMplusYOLO}. In direct QA annotations, the model is supervised only with the final answer to a query, without being required to explicitly reason about whether the task-relevant evidence is visible, small, or distant. As a result, the model may learn annotation shortcuts or dataset-specific response patterns rather than robust visual reasoning grounded in the actual visual evidence. This issue can be further amplified by the imbalanced nature of construction-site monitoring datasets, where positive cases such as safety-rule violations or object-present instances are often much less frequent than negative or object-absent cases. Consequently, the model may become biased toward frequent responses and fail to recognize rare but safety-critical evidence. When the global image resolution is reduced or the target evidence appears small or far from the camera, this shortcut behavior becomes more problematic because ambiguous visual evidence can be incorrectly treated as absent or irrelevant. For example, a distant PPE violation may be predicted as no violation, or a small object may be missed or localized inaccurately. Therefore, direct QA-based adaptation may perform well under controlled, high-resolution settings, but remains vulnerable to long-range monitoring scenarios and reduced-resolution deployment inputs.

\subsubsection{Chain-of-Thought-based Strategies for Generalization}

To mitigate the poor generalization of direct QA-based adaptation, a few studies have explored chain-of-thought (CoT)-based strategies for construction-site VLMs~\cite{ConstructionContext-aware-vlm, ConstructionDoubleThinking, ConstructionEnhancingVLM_VGR_RL}. CoT-based prompting and supervision have been shown to improve generalization in vision-language tasks by encouraging models to perform more explicit and step-by-step visual reasoning~\cite{CoT_for_Generalization1, CoT_for_Generalization2}. Instead of directly mapping an image-level question to the final answer, CoT-based strategies guide the model to generate intermediate reasoning steps before producing the final response. This process can reduce the tendency of direct QA models to rely only on dataset-specific answer patterns, because the model is encouraged to inspect the visual scene more carefully and justify its prediction through an explicit reasoning trace. For example, Chan et al.~\cite{ConstructionContext-aware-vlm} use GroundingDINO~\cite{GroundingDino} to identify the locations of responsible objects, such as PPE items, safety harnesses, excavators, and workers, and augment the training data with object-level grounding information (e.g., bounding box). The model is then fine-tuned to refer to these grounded regions as intermediate evidence during the answer-generation process. By combining step-by-step reasoning with object-level grounding supervision, CoT-based grounding strategies encourage the model to connect relevant visual regions with the final response, thereby improving generalization beyond direct QA-based adaptation. In addition, example image-based CoT can further enhance generalization by providing domain-relevant visual context together with a reasoning-and-answer trace~\cite{MMICL, MultimodalCoT}.

However, these CoT-based strategies still have two important limitations for in-the-wild construction-site monitoring. First, example image-based CoT introduces the additional cost of substantially increased visual-token consumption. Since VLM inference cost is strongly affected by the number of visual tokens, incorporating example images can make inference less efficient. Second, although CoT can encourage more careful and explicit reasoning, it cannot resolve cases where the query-relevant evidence is not sufficiently visible in the input image. When a worker, PPE item, or hazardous object is too small or distant, the model may still be unable to determine whether the target evidence is truly absent or simply insufficiently visible from the global image.

Recent studies in general-domain VLM research have begun to address this perceptual limitation through zoom-in-based visual inspection strategies. Visual CoT~\cite{VisualCoT} introduces region-grounded CoT supervision in which the model identifies query-relevant regions through bounding-box coordinates and subsequently uses the corresponding local visual details to derive the final answer. CropVLM~\cite{CropVLM} improves the efficiency of this localized inspection process by using an external low-cost model to extract local visual details from query-relevant regions and then providing this localized evidence to a large VLM for final answer generation. More recently, AdaptVision~\cite{AdaptVision} further advances this paradigm by enabling adaptive visual acquisition: rather than performing local inspection for every query, the model first reasons over a global image and selectively requests a crop only when additional visual detail is required, thereby allowing either single-turn reasoning from the global view or multi-turn reasoning with local inspection. These studies demonstrate growing interest in coarse-to-fine visual inspection for improving the perception performance of general-purpose VLMs. However, despite these advances in general-domain VLM research, such adaptive zoom-in reasoning has not yet been investigated for construction-site monitoring, where relevant safety evidence is frequently small, distant, or difficult to distinguish in wide-view images. To address this gap, we construct a region-aware CoT dataset that explicitly supervises when and where local inspection is required and use it to train AVA-VLM, thereby adapting adaptive coarse-to-fine visual reasoning to in-the-wild construction-site monitoring.

\subsection{Visual Tokens as the Main Efficiency Bottleneck}

Modern VLM architectures, such as the Qwen~\cite{QwenVL, Qwen2VL, Qwen2_5}, InternVL~\cite{InternVL2_5, InternVL3, InternVL3_5}, and LLaVA~\cite{LLaVA, LLaVA-NeXT, LLaVA-CoT} families, generally consist of three main components: a visual encoder, a modality projector, and a large language model (LLM). The visual encoder converts an input image into a sequence of visual tokens. The modality projector then maps these visual tokens into the embedding space of the LLM, enabling visual features and textual tokens to be jointly processed. Finally, the LLM integrates the aligned visual and textual representations to generate responses.

The computational cost of VLM inference is strongly affected by the input sequence length, which includes system instructions, visual tokens, and question tokens. Among these components, visual tokens often dominate the input sequence and account for a substantial portion of the computational overhead~\cite{VisionThink,AdaptVision}. In typical VLM tasks, prior work reports that the number of visual tokens can exceed the number of system or question tokens by up to 20 times~\cite{AdaptVision}. Therefore, although VLMs are often discussed in terms of model size, the visual-token budget is also a critical factor that determines inference latency and memory consumption.

Since visual tokens dominate the input sequence length, visual-token reduction has become an active research direction in the broader VLM literature~\cite{visualtoken1, visualtoken2, visualtoken3, visualtoken4, visualtoken5, visualtoken6, visualtoken7, VisionThink, AdaptVision}. Existing studies have attempted to reduce redundant visual tokens by pruning tokens based on early-layer attention scores~\cite{visualtoken6}; by progressively compressing tokens across layers to reduce information loss~\cite{visualtoken7}; by selecting query-relevant visual tokens through cross-modal relevance~\cite{visualtoken5}; and by compacting both text-informative and visually redundant tokens~\cite{visualtoken1}. These studies demonstrate that many visual tokens are not equally necessary for generating the final response, suggesting that reducing irrelevant visual information can improve inference efficiency with limited performance degradation.

More recently, RL-based approaches have been explored for adaptive visual-token control. For example, Yang et al.~\cite{VisionThink} trains a model to decide whether a low-resolution image is sufficient or whether the original image is needed, thereby reducing unnecessary high-resolution processing. Lin et al.~\cite{AdaptVision} further improves inference efficiency by learning to determine the minimum amount of visual information required for a given query.

Despite these advances in efficient VLMs, construction-site VLMs have not considered visual-token efficiency as a central design. Existing construction-tailored VLMs still process dense visual representations from the full-size image to maximize task-specific performance, introducing substantial inference overhead~\cite{ConstructionVLM1, ConstructionVLM2, ConstructionVCSQ, ConstructionCS-VLM, ConstructionTran1, ConstructionVLMplusYOLO}. However, such full-image dense processing may be impractical in in-the-wild construction-site monitoring, where camera feeds must often be processed under edge-device compute budgets. In such deployment settings, processing every input as a full-size high-resolution image can be infeasible, even if high-resolution frames are originally available. Moreover, using a full-size image is often unnecessary for construction-site monitoring. In many cases, the query-relevant evidence occupies only a small portion of the image, while large irrelevant regions such as sky, ground, background buildings, or empty areas dominate the visual input. In other cases, the relevant objects are already sufficiently visible from the global view, so a low-resolution image can provide enough information to answer the query without processing the entire image at high resolution.

We address this efficiency gap by combining a low-resolution global image with an adaptively cropped local image. The low-resolution global image provides scene-level context for understanding the overall construction environment, while the cropped image supplies high-resolution visual evidence only for the query-relevant region. In this way, AVA-VLM reduces unnecessary visual tokens from irrelevant areas while preserving detailed information where it is needed. This design makes AVA-VLM more suitable for practical construction-site monitoring, where both accurate visual reasoning and efficient inference are required.

\section{Dataset Curation}
\label{sec:Dataset}

\begin{center}
\includegraphics[
    width=\textwidth,
    height=\textheight,
    keepaspectratio
]{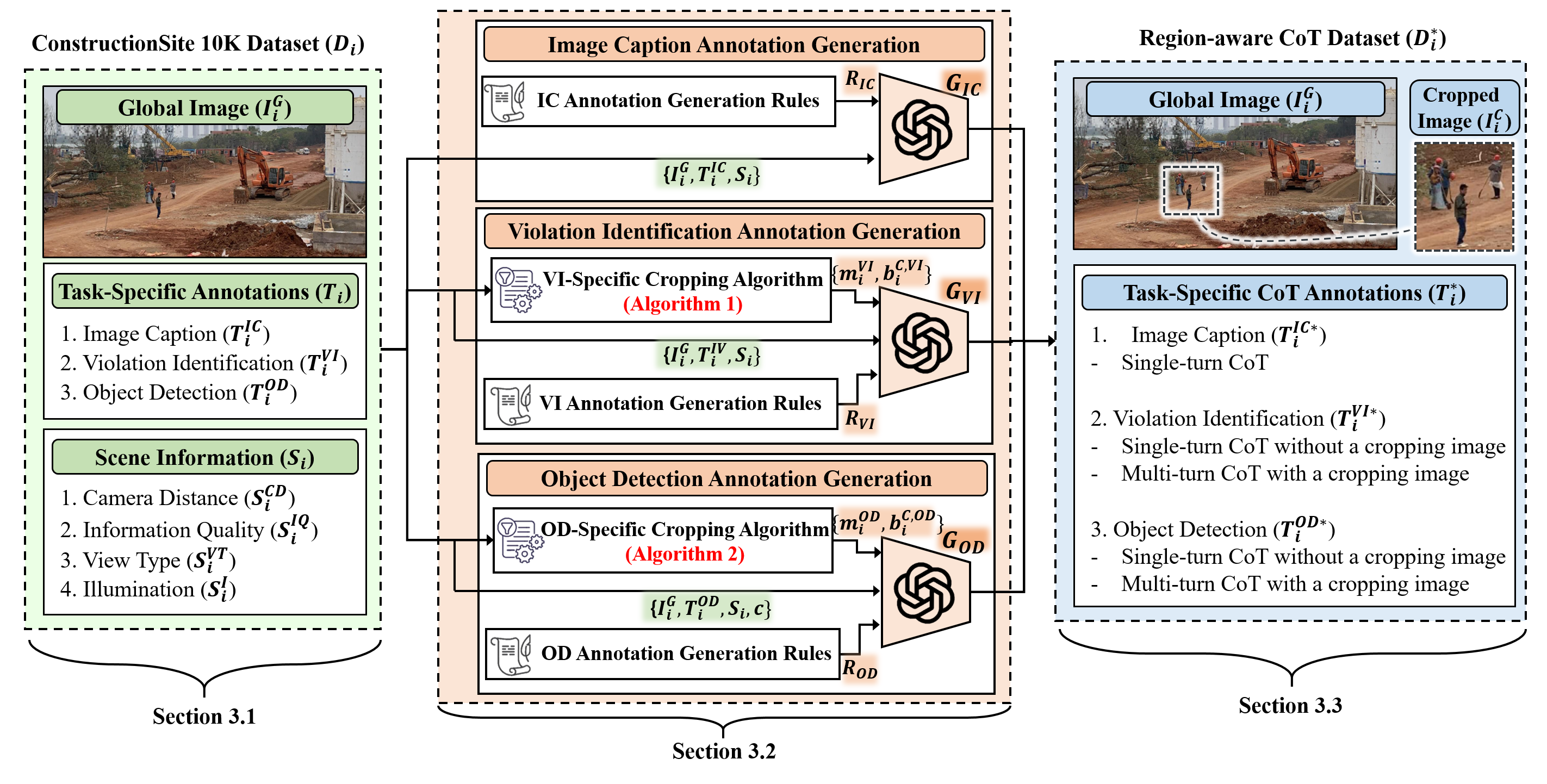}

\captionof{figure}{
Overview of the region-aware CoT dataset curation pipeline. Starting from the source construction-site dataset, we convert task-specific annotations and scene-level information into a region-aware CoT dataset through task-specific annotation generation procedures.
}
\label{fig:DatasetCurationPipeline}
\end{center}

In this section, we describe the dataset curation process used to construct the proposed region-aware CoT dataset. As shown in Figure~\ref{fig:DatasetCurationPipeline}, we start from a construction-site dataset of the direct QA structure~\cite{ConstructionSite10K} that contains global images, task-specific annotations, and scene-level metadata. Based on these original annotations, we generate task-specific CoT annotations for image captioning, violation identification, and object detection. In particular, image captioning is converted into a single-turn CoT format, while violation identification and object detection are converted into adaptive CoT formats that determine whether a query-relevant crop is required and, if so, where the model should inspect. The final curated dataset therefore contains both global scene information and optional cropped local evidence, enabling VLMs to learn region-aware reasoning for construction-site monitoring.

\subsection{Source Data and Annotation Structure}
\label{subsec:SourceData}

We represent the source dataset as $\mathcal{D}=\{D_i\}_{i=1}^{N}$, where each sample $D_i$ contains a global construction-site image, task-specific annotations, and scene-level information. Specifically, each sample is defined as $D_i=\{I_i^{G}, T_i, S_i\}$, where $I_i^{G}$ is the global image, $T_i$ is the set of task-specific annotations, and $S_i$ is the scene-level metadata.

The task-specific annotation set is composed of image captioning (IC), violation identification (VI), and object detection (OD) annotations, i.e., $T_i=\{T_i^{IC},T_i^{VI},T_i^{OD}\}$. Each task annotation consists of a task-specific query and its corresponding ground truth. Specifically, the image captioning annotation is represented as $T_i^{IC}=\{q_i^{IC},y_i^{IC}\}$, where $q_i^{IC}$ denotes the captioning query and $y_i^{IC}$ denotes the ground-truth scene description of the global image. The violation identification annotation is represented as $T_i^{VI}=\{q_i^{VI},y_i^{VI}\}$, where $q_i^{VI}$ denotes the violation-identification query and $y_i^{VI}$ denotes the ground-truth safety-rule annotation. The object detection annotation is represented as $T_i^{OD}=\{q_i^{OD},y_i^{OD}\}$, where $q_i^{OD}$ denotes the object-detection query and $y_i^{OD}$ denotes the ground-truth object annotation.

The ground truth for violation identification, $y_i^{VI}$, describes whether a safety-rule violation is present in the global image and, if so, provides the corresponding violation annotation. Although the original source dataset includes multiple types of construction-safety violations, such as PPE non-compliance, safety-harness violations, missing edge protection, and excavator-proximity hazards, the positive samples are highly imbalanced across categories. In particular, safety-harness and excavator-proximity violations contain fewer than 100 positive instances, while edge-protection violations contain fewer than 200 positive instances. Due to this annotation coverage limitation, it is difficult to reliably train and evaluate all safety-rule categories within the current supervised fine-tuning setting. Since our primary objective is to investigate the operational range of construction-site VLMs in wide-view monitoring scenarios, we focus on PPE non-compliance, which provides 955 positive instances and still requires the model to reason about small or distant workers in challenging construction-site images. Accordingly, the VI task in this study checks whether workers on foot at construction sites properly use PPE, including hard hats, proper clothing covering shoulders and legs, protective shoes that cover the toes, high-visibility retroreflective vests when required, and face shields or safety glasses during cutting, welding, grinding, or drilling. Importantly, this task selection does not restrict the proposed methodology itself to PPE non-compliance. Given sufficient task-specific annotations, the same region-aware CoT annotation generation and adaptive local-inspection framework can be extended to other construction-safety violations.

The global image $I_i^{G}$ may contain multiple workers with distinct PPE conditions. The violation-identification label $c_i^{VI}$ indicates whether PPE non-compliance is present in image $i$, where $c_i^{VI}=1$ denotes the presence of a PPE violation and $c_i^{VI}=0$ denotes no PPE violation. When PPE non-compliance is present, the responsible workers are associated with a set of bounding boxes $\mathcal{B}_i^{VI}=\{b_i^{j}\}_{j=1}^{M_i^{VI}}$, where $M_i^{VI}$ is the number of responsible workers associated with PPE violations in image $i$. Each responsible-worker box is represented as $b_i^{j}=[x_{\min}^{j},y_{\min}^{j},x_{\max}^{j},y_{\max}^{j}]$. To obtain a single violation region, these responsible-worker boxes are merged into one bounding box:
\begin{equation}
b_i^{VI}
=
\left[
\min_{1\leq j\leq M_i^{VI}} x_{\min}^{j},
\min_{1\leq j\leq M_i^{VI}} y_{\min}^{j},
\max_{1\leq j\leq M_i^{VI}} x_{\max}^{j},
\max_{1\leq j\leq M_i^{VI}} y_{\max}^{j}
\right].
\end{equation}
Accordingly, the violation-identification ground truth is defined as
\begin{equation}
y_i^{VI}
=
\begin{cases}
(c_i^{VI}, b_i^{VI}, e_i^{VI}), & \text{if } c_i^{VI}=1, \\
(c_i^{VI}, \varnothing, \varnothing), & \text{if } c_i^{VI}=0,
\end{cases}
\end{equation}
where $b_i^{VI}$ denotes the merged bounding box covering all responsible workers, and $e_i^{VI}$ denotes a textual explanation that summarizes the PPE-violation types observed among the responsible workers.

The ground truth for object detection, $y_i^{OD}$, contains bounding-box annotations for three construction-related object categories:
\begin{equation}
y_i^{OD}=\{O_{i,c}\mid c\in\mathcal{C}^{OD}\}, 
\quad
\mathcal{C}^{OD}=\{\text{Excavator},\text{Rebar},\text{Worker with hard hat}\}.
\end{equation}
For each category $c$, the annotation is represented as
\begin{equation}
O_{i,c} =
\begin{cases}
\{o_{i,c}^{k}\}_{k=1}^{K_{i,c}}, & \text{if objects of category } c \text{ exist}, \\
\varnothing, & \text{otherwise},
\end{cases}
\end{equation}
where $K_{i,c}$ is the number of object instances for category $c$ in image $i$. Each object instance is represented by a bounding box, $o_{i,c}^{k}=b_{i,c}^{k}$, where $b_{i,c}^{k}=[x_{\min},y_{\min},x_{\max},y_{\max}]$ denotes the bounding box of the $k$-th object instance.

In addition to task-specific annotations, each sample includes scene-level information, 
represented as $S_i=\{S_i^{CD},S_i^{IQ},S_i^{VT},S_i^{I}\}$, where each superscript denotes camera distance, information quality, view type, and illumination, respectively, $S_i^{CD} \in \{\text{Short}, \text{Mid}, \text{Long}\}$, 
$S_i^{IQ} \in \{\text{Sparse}, \text{Rich}\}$, $S_i^{VT} \in \{\text{Elevation view}, \text{Plan view}\}$, and $S_i^{I}  \in \{\text{Normal}, \text{Night}, \text{Overexposed}, \text{Underexposed}\}$. Thus, overall, each sample can be summarized as
\begin{equation}
D_i =
\left\{
I_i^{G},
\{T_i^{IC},T_i^{VI},T_i^{OD}\},
\{S_i^{CD},S_i^{IQ},S_i^{VT},S_i^{I}\}
\right\}.
\end{equation}
This structured representation serves as the input to the region-aware CoT annotation generation process described in the following subsection.

\subsection{Region-Aware CoT Annotation Generation}
\label{subsec:AnnotationGeneration}

Based on the source data structure defined in Section~\ref{subsec:SourceData}, we generate task-specific region-aware CoT annotations for image captioning, violation identification, and object detection. The goal of this process is to convert the original direct-QA annotations into response traces that explicitly include intermediate reasoning steps and, when necessary, query-relevant cropped visual evidence. As illustrated in Figure~\ref{fig:DatasetCurationPipeline}, we adapt GPT-4o as task-specific generators, denoted as $G_{IC}$, $G_{VI}$, and $G_{OD}$, to generate CoT annotations for image captioning, violation identification, and object detection, respectively.

For each task, the generator receives the global image $I_i^{G}$, the corresponding task annotation $T_i^{(\cdot)}$, scene-level information $S_i$, and task-specific annotation generation rules. We denote the annotation generation rules for image captioning, violation identification, and object detection as $\mathcal{R}_{IC}$, $\mathcal{R}_{VI}$, and $\mathcal{R}_{OD}$, respectively. These rules specify the required response format, the use of \texttt{<think>} and \texttt{<answer>} tags, and, for adaptive tasks, the condition under which a local crop should be requested through a tool call. 

For image captioning, no cropping operation is used because the task aims to summarize the overall scene context rather than inspect a specific local region. Therefore, the captioning generator produces a CoT-style target response $y_i^{IC*}$ from the original captioning ground truth $y_i^{IC}$:
\begin{equation}
y_i^{IC*}
=
G_{IC}\left(I_i^{G}, T_i^{IC}, S_i;\mathcal{R}_{IC}\right).
\end{equation}
The resulting captioning CoT annotation is represented as $T_i^{IC*}=\{q_i^{IC},y_i^{IC*}\}$, where $y_i^{IC*}$ follows the single-turn format \texttt{<think>} $\rightarrow$ \texttt{<answer>}. The generated reasoning briefly describes the visible scene elements before producing the final caption $y_i^{IC}$ inside the answer.

For violation identification and object detection, the annotation generation process is adaptive. For efficient inference, the model should answer directly when the relevant evidence is sufficiently visible from the global image, but it should request a local crop when the target evidence is small, distant, or difficult to inspect. Therefore, before generating CoT annotations for these tasks, we first determine whether a crop is required using task-specific cropping algorithms.

For violation identification, positive PPE-violation samples use the merged responsible-worker box $b_i^{VI}$ in $y_i^{VI}$ to determine whether the violation evidence is too small in the global image. For no-violation samples, $y_i^{VI}$ does not contain a violation-region bounding box. To still expose the model to realistic no-violation cases that require close inspection, we use an auxiliary YOLO-based detector for mid- and long-distance no-violation images. Specifically, we use a YOLO26-large model finetuned on the Construction-Hazard-Detection Computer Vision Dataset\footnote{\url{https://universe.roboflow.com/object-detection-qn97p/construction-hazard-detection}} to extract construction-related visual cues, including workers, no-vest, vest, no-helmet, and helmet. We denote this detector as $D_{\text{YOLO}}$. Given a global image, it produces a set of detections $\mathcal{Z}_i = D_{\text{YOLO}}(I_i^{G})$, where each detection contains a class label and bounding box. For no-violation samples, detected small worker boxes are used as candidate regions for detailed inspection, and PPE-related detections inside the worker boxes are used to guide the generated reasoning while keeping the final answer as no violation.

\begin{algorithm}[h]
\caption{VI-Specific Cropping Algorithm}
\label{alg:vi_crop}
\begin{algorithmic}[1]
\Require Global image $I_i^{G}$, camera distance $S_i^{CD}$, VI ground truth $y_i^{VI}$, object-level area threshold $\tau_{VI}^{obj}$, crop-region area threshold $\tau_{VI}^{crop}$, expansion ratio $\alpha$, YOLO detector $D_{\text{YOLO}}$
\Ensure Crop indicator $m_i^{VI}$ and crop bounding box $b_i^{C,VI}$
\State Initialize $m_i^{VI}\leftarrow 0$ and $b_i^{C,VI}\leftarrow \varnothing$
\If{$S_i^{CD}\notin\{\text{Mid},\text{Long}\}$}
    \State \Return $m_i^{VI}, b_i^{C,VI}$
\EndIf
\If{$c_i^{VI}=1$}
    \State Extract the merged responsible-worker box $b_i^{VI}$ from $y_i^{VI}$
    \State Set $\mathcal{B}_i \leftarrow \{b_i^{VI}\}$
\Else
    \State Run $D_{\text{YOLO}}$ on $I_i^{G}$ to obtain detections $\mathcal{Z}_i$
    \State Extract detected worker boxes $\mathcal{B}_i^{W}$ and PPE-related detections from $\mathcal{Z}_i$
    \If{$\mathcal{B}_i^{W}=\varnothing$}
        \State \Return $m_i^{VI}, b_i^{C,VI}$
    \EndIf
    \State Set $\mathcal{B}_i \leftarrow \mathcal{B}_i^{W}$
\EndIf
\State Select small boxes $\mathcal{B}_i^{S}=\{b\in\mathcal{B}_i \mid \mathrm{Area}(b)/\mathrm{Area}(I_i^{G}) < \tau_{VI}^{obj}\}$
\If{$\mathcal{B}_i^{S}=\varnothing$}
    \State \Return $m_i^{VI}, b_i^{C,VI}$
\EndIf
\State Compute the union box $b_i^{U}=\mathrm{Union}(\mathcal{B}_i^{S})$
\If{$\mathrm{Area}(b_i^{U})/\mathrm{Area}(I_i^{G}) > \tau_{VI}^{crop}$}
    \State \Return $m_i^{VI}, b_i^{C,VI}$
\EndIf
\State Expand and clip the crop box: $b_i^{C,VI}=\mathrm{Expand}(b_i^{U},\alpha,I_i^{G})$
\State Set $m_i^{VI}\leftarrow 1$
\State \Return $m_i^{VI}, b_i^{C,VI}$
\end{algorithmic}
\end{algorithm}

Algorithm~\ref{alg:vi_crop} first checks whether the image is captured from a mid- or long-distance viewpoint, because short-distance images typically provide sufficient visual detail from the global view. If PPE non-compliance is present, the algorithm uses the merged responsible-worker box $b_i^{VI}$ from $y_i^{VI}$ as the candidate crop region. If no PPE violation is present, the algorithm instead uses $D_{\text{YOLO}}$ to identify small workers and PPE-related cues. Candidate boxes whose area ratios exceed the threshold $\tau_{VI}^{obj}$ are treated as sufficiently visible and excluded from cropping. The remaining small boxes are merged and expanded to form a crop region for detailed inspection.

After obtaining the VI crop decision, the violation-identification generator produces a CoT-style target response $y_i^{VI*}$ from the original VI ground truth $y_i^{VI}$:
\begin{equation}
y_i^{VI*}
=
G_{VI}\left(I_i^{G}, T_i^{VI}, S_i, m_i^{VI}, b_i^{C, VI};\mathcal{R}_{VI}\right),
\end{equation}
where $m_i^{VI}\in\{0,1\}$ indicates whether the crop is used. The resulting VI CoT annotation is represented as $T_i^{VI*}=\{q_i^{VI},y_i^{VI*}\}$. If $m_i^{VI}=0$, $y_i^{VI*}$ is generated as a single-turn CoT response in the format \texttt{<think>} $\rightarrow$ \texttt{<answer>}. If $m_i^{VI}=1$, $y_i^{VI*}$ is generated as a multi-turn CoT response in the format \texttt{<think>} $\rightarrow$ \texttt{<tool\_call>} $\rightarrow$ \texttt{<think>} $\rightarrow$ \texttt{<answer>}, where the second reasoning step incorporates the cropped image as local visual evidence. In both cases, the final answer inside $y_i^{VI*}$ is constrained to match the image-level violation label $c_i^{VI}$ and, when $c_i^{VI}=1$, the aggregated violation explanation $e_i^{VI}$ in $y_i^{VI}$.

For object detection, the crop decision is made separately for each target object category $c\in\mathcal{C}^{OD}$. Unlike no-violation VI samples, OD annotations already provide object-level bounding boxes in $y_i^{OD}$, so the cropping decision can be made directly from the target-object boxes.

\begin{algorithm}[h]
\caption{OD-Specific Cropping Algorithm}
\label{alg:od_crop}
\begin{algorithmic}[1]
\Require Global image $I_i^{G}$, camera distance $S_i^{CD}$, OD ground truth $y_i^{OD}$, target object category $c$, object-level area threshold $\tau_{OD}^{obj}$, crop-region area threshold $\tau_{OD}^{crop}$, expansion ratio $\alpha$
\Ensure Crop indicator $m_{i,c}^{OD}$ and crop bounding box $b_{i,c}^{C,OD}$
\State Initialize $m_{i,c}^{OD}\leftarrow 0$ and $b_{i,c}^{C,OD}\leftarrow \varnothing$
\If{$S_i^{CD}\notin\{\text{Mid},\text{Long}\}$}
    \State \Return $m_{i,c}^{OD}, b_{i,c}^{C,OD}$
\EndIf
\State Extract target-object boxes $O_{i,c}$ from $y_i^{OD}$
\If{$O_{i,c}=\varnothing$}
    \State \Return $m_{i,c}^{OD}, b_{i,c}^{C,OD}$
\EndIf
\State Select small target boxes $\mathcal{B}_{i,c}^{S}=\{b\in O_{i,c} \mid \mathrm{Area}(b)/\mathrm{Area}(I_i^{G}) < \tau_{OD}^{obj}\}$
\If{$\mathcal{B}_{i,c}^{S}=\varnothing$}
    \State \Return $m_{i,c}^{OD}, b_{i,c}^{C,OD}$
\EndIf
\State Compute the union box $b_{i,c}^{U}=\mathrm{Union}(\mathcal{B}_{i,c}^{S})$
\If{$\mathrm{Area}(b_{i,c}^{U})/\mathrm{Area}(I_i^{G}) \geq \tau_{OD}^{crop}$}
    \State \Return $m_{i,c}^{OD}, b_{i,c}^{C,OD}$
\EndIf
\State Expand and clip the crop box: $b_{i,c}^{C,OD}=\mathrm{Expand}(b_{i,c}^{U},\alpha,I_i^{G})$
\State Set $m_{i,c}^{OD}\leftarrow 1$
\State \Return $m_{i,c}^{OD}, b_{i,c}^{C,OD}$
\end{algorithmic}
\end{algorithm}

Algorithm~\ref{alg:od_crop} follows the same coarse-to-fine principle but uses object-level ground-truth annotations from $y_i^{OD}$. For each target object category, the algorithm determines whether the object exists, filters out boxes that are already sufficiently large in the global image, and generates an expanded crop around the remaining small target boxes.

After obtaining the OD crop decision, the object-detection generator produces a CoT-style target response $y_{i,c}^{OD*}$ for each object category $c\in\mathcal{C}^{OD}$:
\begin{equation}
y_{i,c}^{OD*}
=
G_{OD}\left(I_i^{G}, T_i^{OD}, S_i, c, m_{i,c}^{OD}, b_{i,c}^{C, OD};\mathcal{R}_{OD}\right).
\end{equation}
The resulting OD CoT annotation is represented as $T_{i,c}^{OD}=\{q_{i,c}^{OD},y_{i,c}^{OD*}\}$, where $q_{i,c}^{OD}$ denotes the object-specific detection query for category $c$. If $m_{i,c}^{OD}=0$, $y_{i,c}^{OD*}$ directly reasons from the global image and outputs the object bounding boxes in $y_i^{OD}$. If $m_{i,c}^{OD}=1$, $y_{i,c}^{OD*}$ first requests a query-relevant crop and then reasons over the cropped image before producing the final detection answer. In both cases, the final answer inside $y_{i,c}^{OD*}$ is constrained to match the corresponding object annotation in $y_i^{OD}$.

Based on the generated task-specific CoT annotations, the final region-aware CoT sample is represented as
\begin{equation}
D_i^{*}
=
\left\{
I_i^{G},
I_i^{C},
T_i^{*}\
\right\},
\end{equation}
where $I_i^{G}$ denotes the global image, $I_i^{C}$ denotes the optional cropped image generated by the crop tool, and $T_i^{*}
=
\left\{
T_i^{IC*},
T_i^{VI*},
\{T_{i,c}^{OD*}\}_{c\in\mathcal{C}^{OD}}
\right\}$ denotes the task-specific CoT annotation set.
Overall, each task-specific generator produces the final CoT annotation after the corresponding crop decision is determined. If no crop is required, the response contains a single reasoning step followed by the answer. If a crop is required, the response first explains why the local region should be inspected, calls the crop tool with $b_i^{C, VI}$ or $b_{i,c}^{C, OD}$, and then uses the cropped image to complete the reasoning before generating the final answer. This process yields task-specific CoT annotations $T_i^{IC*}$, $T_i^{VI*}$, and $T_{i,c}^{OD*}$ that explicitly connect global context, local visual evidence, and final task responses.

To qualitatively illustrate the generated annotations, Figure~\ref{fig:example} presents representative examples from the region-aware CoT dataset across image captioning, violation identification, and object detection tasks. For image captioning, the annotation follows a non-zoom-in structure because the task aims to describe the overall construction scene, including people, equipment, materials, activities, and spatial relationships, rather than to inspect a specific query-relevant local region. For violation identification and object detection, the examples include both zoom-in and non-zoom-in cases. This design is important because the model should not always request a local crop; instead, it should learn to adaptively decide whether detailed inspection is necessary based on the visibility of query-relevant evidence in the global image. When workers, PPE conditions, or target objects are small, distant, or visually ambiguous, the generated annotation includes a crop request followed by reasoning over the cropped image. In contrast, when the relevant evidence is already sufficiently visible, the generated annotation directly answers from the global image without using the crop tool.

\begin{center}
\includegraphics[
    width=0.99\textwidth,
    height=0.99\textheight,
    keepaspectratio
]{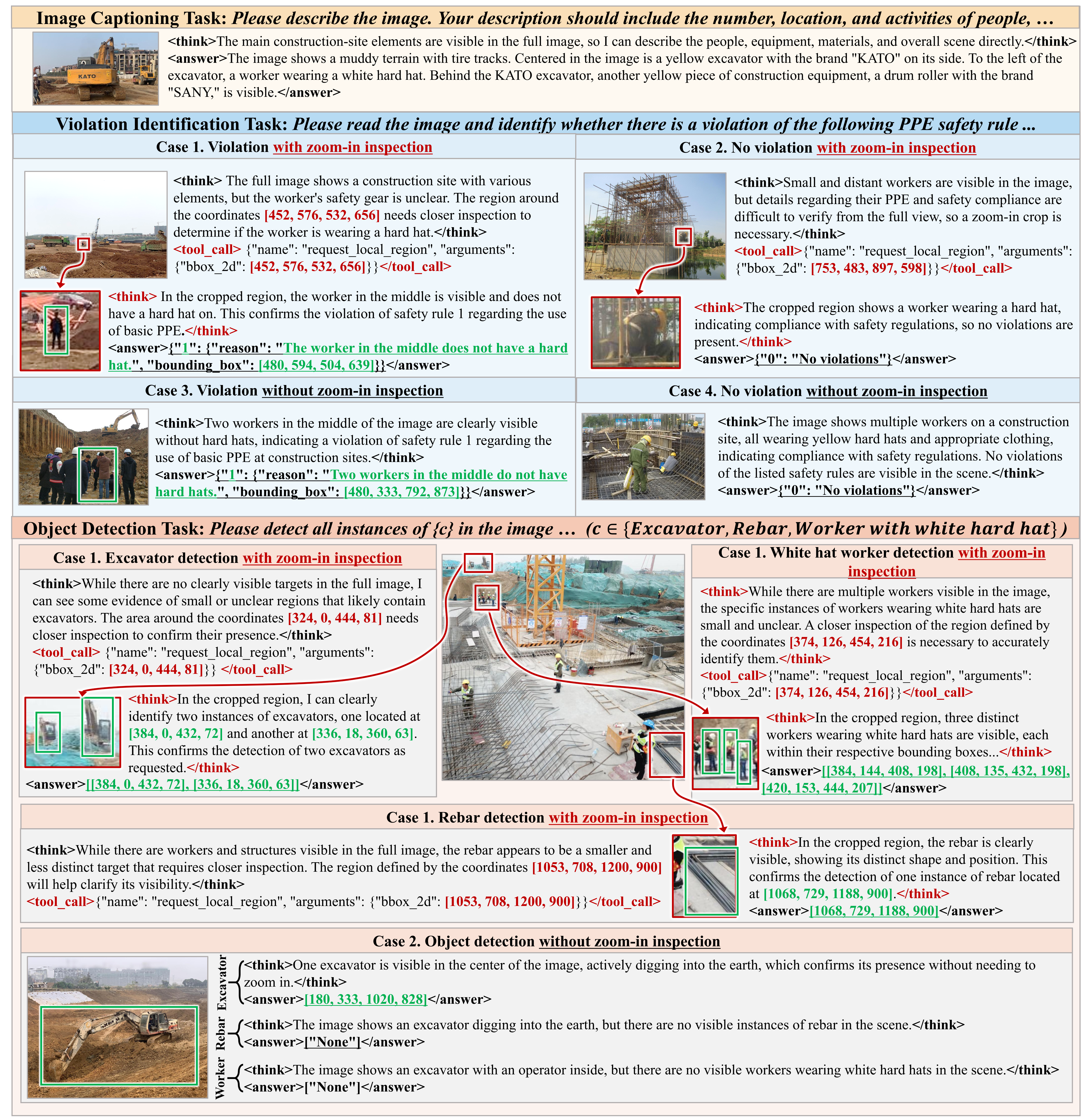}

\captionof{figure}{
Examples of the region-aware CoT dataset.
}
\label{fig:example}
\end{center}

 \subsection{Region-Aware CoT Dataset Statistics}
\label{subsec:RA-CoT-Dataset}

We summarize the statistics of the constructed region-aware CoT dataset. Since the source dataset does not provide a separate validation split, we first divide the original training split into training and validation sets with an 8:2 ratio. During this split, we preserve the balance of key dataset properties, including camera distance categories (short, mid, and long), crop-tool usage, and the number of task-specific instances in the VI and OD tasks. The original test split is kept unchanged and used only for evaluation.

\begin{table}[t]
\centering
\caption{General dataset statistics for train, validation, and test splits.}
\label{tab:dataset_statistics}
\begin{tabular}{llrrrr}
\toprule
Task & Statistic & Train & Val & Test & Total \\
\midrule
\multirow{2}{*}{Image Captioning}
& \# Samples & 5,509 & 1,500 & 3,004 & 10,013 \\
& \# Unique images & 5,509 & 1,500 & 3,004 & 10,013 \\
\midrule
\multirow{4}{*}{Violation Identification}
& \# Samples & 5,341 & 1,455 & 2,896 & 9,692 \\
& \# No-violation samples & 4,814 & 1,330 & 2,593 & 8,737 \\
& \# Violation samples & 527 & 125 & 303 & 955 \\
\midrule
\multirow{5}{*}{Object Detection}
& \# Samples & 16,527 & 4,500 & 9,012 & 30,039 \\
& \# Total instances & 5,120 & 1,567 & 2,373 & 9,060 \\
& \# Excavator instances & 2,458 & 809 & 1,414 & 4,681 \\
& \# Rebar instances & 1,869 & 542 & 567 & 2,978 \\
& \# Worker-with-white-hard-hat instances & 793 & 216 & 392 & 1,401 \\
\bottomrule
\end{tabular}
\end{table}

Table~\ref{tab:dataset_statistics} shows the general statistics of the curated dataset across the training, validation, and test splits. For each source image, we generate one image-captioning annotation, one violation-identification annotation, and object-detection annotations for the target object categories. The IC task contains 10,013 samples in total, corresponding to the number of unique images. The VI task contains 9,692 valid samples after excluding unavailable or invalid annotations, among which 955 samples contain violations and 8,737 samples correspond to no-violation cases. The OD task contains 30,039 samples because one image can produce multiple object-specific detection queries. Across all splits, the OD annotations include 4,681 excavator instances, 2,978 rebar instances, and 1,401 worker-with-white-hard-hat instances.

\begin{table}[t]
\centering
\caption{Image-level distance statistics and cropping distribution.}
\label{tab:distance_image_statistics}
\begin{tabularx}{\textwidth}{Xrrrrrrr}
\toprule
Distance 
& Train images & Train crop 
& Val images & Val crop 
& Test images 
& Train+Val images 
& Crop ratio (\%) \\
\midrule
Short & 1,315 & 0 & 245 & 0 & 1,360 & 1,560 & 0.0 \\
Mid   & 3,924 & 2,956 & 1,139 & 812 & 1,310 & 5,063 & 74.4 \\
Long  & 270 & 210 & 116 & 84 & 334 & 386 & 76.2 \\
\midrule
Total & 5,509 & 3,166 & 1,500 & 896 & 3,004 & 7,009 & 58.0 \\
\bottomrule
\end{tabularx}
\end{table}

Table~\ref{tab:distance_image_statistics} presents image-level distance statistics and crop-tool usage. An image is counted as requiring a crop if at least one of its generated VI or OD annotations uses the crop tool. Specifically, for image $i$, we define the image-level crop indicator as
\begin{equation}
m_i^{\mathrm{img}}
=
\max\left(
m_i^{VI},
\max_{c\in\mathcal{C}^{OD}} m_{i,c}^{OD}
\right),
\end{equation}
where $m_i^{VI}\in\{0,1\}$ and $m_{i,c}^{OD}\in\{0,1\}$ indicate whether the VI annotation and the OD annotation for object category $c$ use a crop, respectively.

At the image level, crop-tool usage is frequent in mid- and long-distance images, with crop ratios of 74.4\% and 76.2\% in the training and validation splits, respectively. This trend reflects the wide-view nature of the dataset, where query-relevant evidence can appear small, distant, or visually ambiguous in the global image. The slightly higher crop ratio for long-distance images is also expected because local evidence becomes harder to inspect as camera distance increases. Importantly, these values should be interpreted at the image level: since each image contains multiple task annotations, an image is counted as crop-required if any VI or OD annotation requires local inspection. Therefore, the high crop ratios indicate that many wide-view images contain at least one query-relevant region that benefits from zoom-in inspection, rather than implying that every task annotation within those images requires cropping.

\begin{table}[t]
\centering
\caption{Annotation-level distance statistics and cropping distribution.}
\label{tab:distance_annotation_statistics}
\begin{tabularx}{\textwidth}{Xrrrrrrr}
\toprule
Distance 
& Train ann. & Train crop 
& Val ann. & Val crop 
& Test ann. 
& Train+Val ann. 
& Crop ratio (\%) \\
\midrule
Short & 6,552 & 0 & 1,222 & 0 & 6,767 & 7,774 & 0.0 \\
Mid   & 19,482 & 4,123 & 5,654 & 1,104 & 6,487 & 25,136 & 20.8 \\
Long  & 1,343 & 307 & 579 & 123 & 1,658 & 1,922 & 22.4 \\
\midrule
Total & 27,377 & 4,430 & 7,455 & 1,227 & 14,912 & 34,832 & 16.2 \\
\bottomrule
\end{tabularx}
\end{table}

Table~\ref{tab:distance_annotation_statistics} reports the distance statistics and crop-tool usage at the annotation level. Unlike the image-level analysis, the annotation-level crop ratio is computed separately for each generated task annotation. Therefore, it measures how often the crop tool is actually used for individual task queries rather than whether an image contains at least one crop-required task. Since image captioning does not use cropping, we compute the annotation-level crop ratio over the adaptive VI and OD annotations as

\begin{equation}
\rho^{\mathrm{ann}}
=
\frac{
\sum_i m_i^{VI} + \sum_i\sum_{c\in\mathcal{C}^{OD}} m_{i,c}^{OD}
}{
N^{VI} + \sum_{c\in\mathcal{C}^{OD}} N_c^{OD}
},
\end{equation}
where $m_i^{VI}$ and $m_{i,c}^{OD}$ denote the crop indicators for the VI annotation and the OD annotation of category $c$, respectively. $N^{VI}$ denotes the number of VI annotations, and $N_c^{OD}$ denotes the number of OD annotations for category $c$.

At the annotation level, the crop ratios are much lower than the image-level crop ratios. In the training and validation splits, the overall annotation-level crop ratio is 16.2\%, with 20.8\% for mid-distance annotations and 22.4\% for long-distance annotations. This result highlights the adaptive nature of the proposed dataset. Even when an image is captured from a wide monitoring viewpoint, not every query requires high-resolution local inspection. If the target evidence is already sufficiently visible from the global image, the annotation remains a single-turn CoT response without a tool call, as shown in the non-zoom-in examples of the VI (case 3 and 4) and OD (case 2) tasks in Figure~\ref{fig:example}. Cropping is introduced only when the query-relevant evidence is small, distant, or visually ambiguous. This design encourages the model to use the crop tool selectively, preserving global-scene reasoning while allocating additional visual attention only to regions that require detailed inspection.

Overall, these statistics show that the curated dataset supports both global reasoning and adaptive local inspection. The high image-level crop ratio of 58.0\% confirms that many construction-site images, especially those captured from mid- and long-distance viewpoints, contain at least one query-relevant region that can benefit from zoom-in inspection. In contrast, the relatively low annotation-level crop ratio of 16.2\% shows that crop-tool usage remains selective at the individual task-query level, allowing AVA-VLM to infer more efficiently.

\section{Methodology}
\label{sec:Methodology}

In this section, we introduce AVA-VLM, an adaptive visual attention framework for practical construction-site monitoring. The key idea of AVA-VLM is to train a VLM to decide whether global visual evidence is sufficient for a given query or whether additional local inspection is required. Instead of always processing the entire image at a high resolution, AVA-VLM learns a human-like coarse-to-fine visual attention behavior: it first reasons over the global image and selectively requests a query-relevant local crop only when the target evidence is visually ambiguous. This design enables the model to preserve global scene understanding while improving robustness across different camera distances and reducing unnecessary visual-token computation during inference.

Given the region-aware CoT dataset $\mathcal{D}^{*}$ constructed in Section~\ref{sec:Dataset}, each training sample contains a global image $I_i^{G}$, an optional cropped image $I_i^{C}$, and task-specific CoT annotations $T^{*}_{i}$. In $T^{*}_{i}$, the crop indicator is denoted by $m_i^{(\cdot)} \in {0,1}$, where $m_i^{(\cdot)}=1$ indicates that local inspection is required and $m_i^{(\cdot)}=0$ indicates that the model should directly answer from the global image. In the following subsections, we describe the training and inference pipelines of AVA-VLM.

\subsection{Training Phase}
\label{subsec:Training}

The goal of the training phase is to teach AVA-VLM two complementary behaviors: (1) adaptive cropping behavior, in which the model learns when and where to request a zoom-in inspection, and (2) final answer generation behavior, in which the model produces the task-specific answer either directly from the global image or after observing the cropped local evidence. The overall framework for AVA-VLM training is illustrated in Figure~\ref{fig:framework} (Top). During training, we use the full-size global image to fully expose the model to the available visual information. We define the trainable AVA-VLM as $f_{\theta_0,\phi}$, where $\theta_0$ represents the pretrained VLM parameters and $\phi$ represents the LoRA parameters. We keep the pretrained VLM weights fixed and update only the LoRA parameters for efficient construction-site adaptation.

Training is performed using the ground-truth response sequences provided in the region-aware CoT dataset $\mathcal{D}^{*}$. Importantly, during training, the crop tool is not executed based on the model's predicted crop region. Instead, each training sample already specifies whether local inspection is required and, if so, which ground-truth cropped image should be used. This design is important because an incorrect predicted crop would provide irrelevant or incomplete visual evidence to the second response, making the answer-generation supervision noisy. By using the ground-truth crop, the model can learn how to interpret the correct local evidence through the final answer loss, while the tool call loss separately teaches the model how to request the appropriate local region. This decoupling stabilizes training by preventing crop-selection errors from directly corrupting the supervision of the final answer.

\begin{center}
\includegraphics[
    width=\textwidth,
    height=\textheight,
    keepaspectratio
]{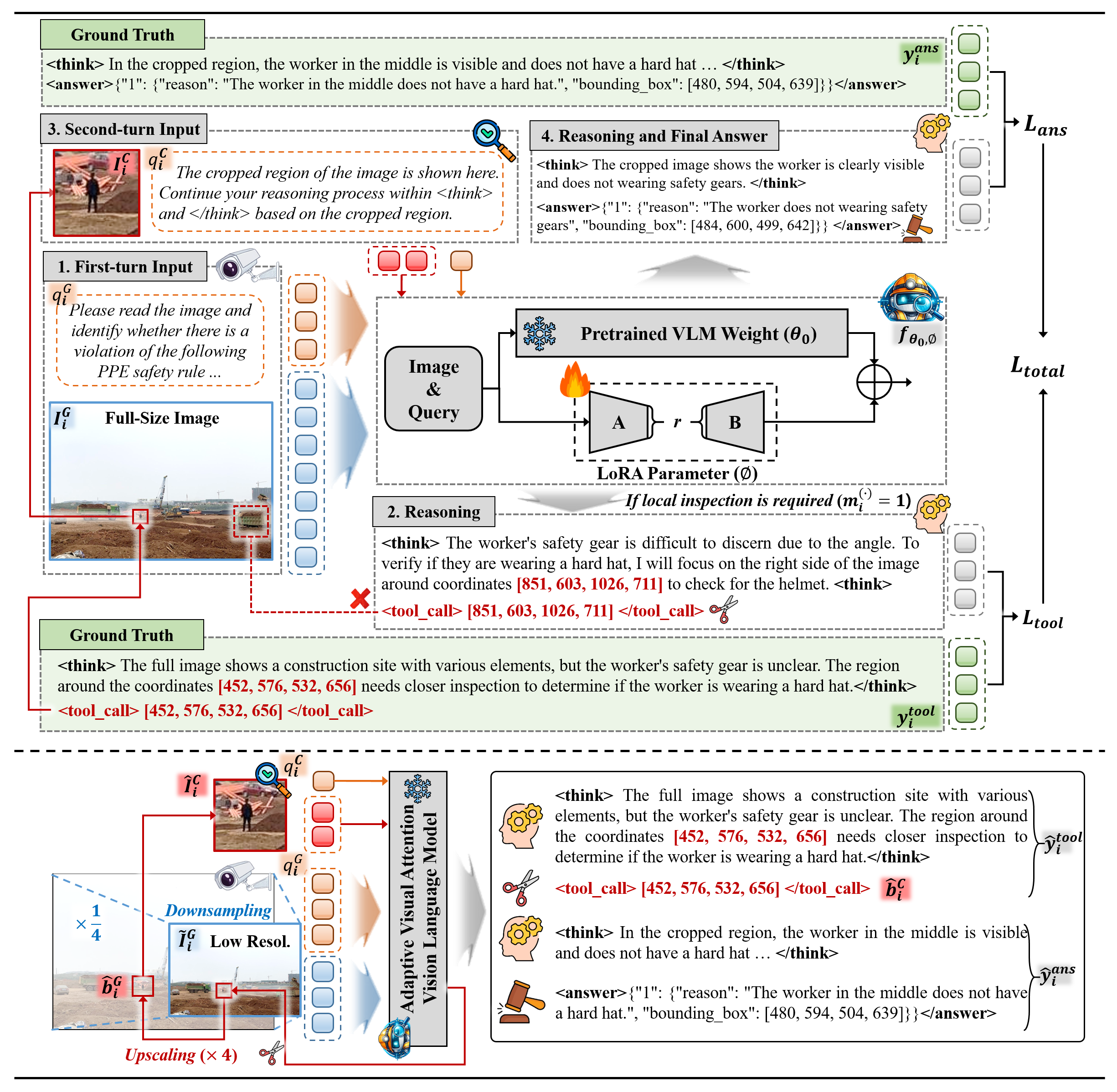}

\captionof{figure}{
Framework of AVA-VLM.
\textbf{Top:} Training pipeline, where AVA-VLM is trained with ground-truth response traces for both direct-answer and crop-required samples. For crop-required samples, the model first learns to generate a local-inspection request and then learns to produce the final answer using the ground-truth cropped image, which decouples crop-region learning from crop-conditioned answer generation.
\textbf{Bottom:} Inference pipeline, where AVA-VLM first reasons over a downsampled global image for efficient visual-token usage and, when needed, predicts a crop region that is mapped back to the original image for detailed inspection.
}
\label{fig:framework}
\end{center}

If no local inspection is required (i.e., $m_i^{(\cdot)}=0$), the model is trained to generate a single final response from the global image and the task query:
\[
I_i^{G}, q_i^G
\rightarrow
\texttt{\textless think\textgreater}
\rightarrow
\texttt{\textless answer\textgreater}.
\]
This case teaches the model to directly answer when the global image already provides sufficient visual evidence.

If local inspection is required (i.e., $m_i^{(\cdot)}=1$), the model is trained with two response targets. First, given the global image and the task query, the model is supervised to generate a local-inspection request:
\[
I_i^{G}, q_i^G
\rightarrow
\texttt{\textless think\textgreater}
\rightarrow
\texttt{\textless tool\_call\textgreater}.
\]
Second, the ground-truth cropped image $I_i^{C}$ and the crop-inspection query $q_i^{C}$ are provided as additional input, and the model is supervised to generate the final crop-conditioned response:
\[
I_i^{G}, q_i^G,
\texttt{\textless think\textgreater}
\rightarrow
\texttt{\textless tool\_call\textgreater},
I_i^{C}, q_i^{C}
\rightarrow
\texttt{\textless think\textgreater}
\rightarrow
\texttt{\textless answer\textgreater}.
\]
Thus, for crop-required samples, the first target teaches the model when and where to request local inspection, while the second target teaches the model how to answer after observing the correct local evidence.

We define two types of ground-truth assistant response targets. First, $\mathbf{y}_{i}^{tool}$ denotes the local-inspection request response target, which includes both the reasoning tokens and the tool-call tokens:
\[
\mathbf{y}_{i}^{tool}
=
\left[
\texttt{\textless think\textgreater}, \ldots, \texttt{\textless /think\textgreater},
\texttt{\textless tool\_call\textgreater}, \ldots, \texttt{\textless /tool\_call\textgreater}
\right].
\]
The corresponding loss is denoted as $\mathcal{L}_{tool}$. Second, $\mathbf{y}_{i}^{ans}$ denotes the final reasoning-and-answer response target:
\[
\mathbf{y}_{i}^{ans}
=
\left[
\texttt{\textless think\textgreater}, \ldots, \texttt{\textless /think\textgreater},
\texttt{\textless answer\textgreater}, \ldots, \texttt{\textless /answer\textgreater}
\right].
\]
The corresponding loss is denoted as $\mathcal{L}_{ans}$. Here, $\mathcal{L}_{ans}$ supervises the entire final assistant response, including both the reasoning step and the final answer, rather than only the tokens inside the \texttt{\textless answer\textgreater} tag.

For each assistant response target, we denote the multimodal context by $\mathbf{x}$ and the corresponding ground-truth assistant response by $\mathbf{y}=(y_1,\ldots,y_T)$. The context $\mathbf{x}$ includes all tokens and visual inputs that precede the target assistant response, such as the system prompt, user query, global image, and, when applicable, the tool-use request, cropped image, and crop-inspection query. The model is optimized to maximize the likelihood of the ground-truth assistant response tokens conditioned on this context. The assistant-token negative log-likelihood is defined as
\begin{equation}
\mathcal{L}(\mathbf{y}\mid\mathbf{x})
=
-\sum_{t=1}^{T}
\log p_{\theta_0,\phi}
\left(
y_t
\mid
\mathbf{x}, y_{<t}
\right).
\end{equation}

For samples without a ground-truth tool call, the model is supervised only with the final reasoning-and-answer target $\mathbf{y}_{i}^{ans}$. For samples with a ground-truth tool call, the model is supervised with both the local-inspection request target $\mathbf{y}_{i}^{tool}$ and the final crop-conditioned answer target $\mathbf{y}_{i}^{ans}$. Therefore, the overall training loss is written as
\begin{equation}
\mathcal{L}_{total}
=
\mathcal{L}_{ans}
+
\mathbb{1}\left[m_i^{(\cdot)}=1\right]\mathcal{L}_{tool}.
\end{equation}
Only assistant-generated tokens are included in the loss. System prompts, user queries, and image tokens are used as context but are masked out during loss computation. Algorithm~\ref{alg} summarizes the training procedure.

\subsection{Inference Phase}
\label{subsec:Inference}

The inference follows the same decision structure as training, but uses a downsampled global image as the first input for efficient inference. During training, AVA-VLM learns two types of behavior: directly answering from the global image when the visual evidence is sufficient, and requesting local inspection when additional visual detail is needed. Based on this, we reduce redundant visual-token computation at inference time by replacing the full-size global image with a downsampled version while preserving access to high-resolution local evidence through the crop tool.

\begin{algorithm}[h]
\caption{Masked supervised training of AVA-VLM}\label{alg}
\begin{algorithmic}[1]
\Require Region-aware CoT dataset $\mathcal{D}^{*}$, pretrained VLM parameters $\theta_0$, LoRA parameters $\phi$
\Ensure Tuned AVA-VLM $f_{\theta_0,\phi}$

\State Freeze pretrained VLM parameters $\theta_0$ and initialize trainable LoRA parameters $\phi$

\For{each training sample $D_i^{}\in\mathcal{D}^{}$}

\If{$m_i^{(\cdot)}=0$}
    \State Construct the answer context $\mathbf{x}_{i}^{ans}$ using the global image $I_i^{G}$ and task query $q_i^G$
    \State Set the ground-truth final response target $\mathbf{y}_{i}^{ans}=\texttt{\textless think\textgreater}\rightarrow\texttt{\textless answer\textgreater}$
    \State Form the supervised sequence $\mathbf{s}_i=[\mathbf{x}_{i}^{ans},\mathbf{y}_{i}^{ans}]$
    \State Unmask only the assistant tokens in $\mathbf{y}_{i}^{ans}$
    \State Set $\mathcal{L}_{total}\leftarrow \mathcal{L}(\mathbf{y}_{i}^{ans}\mid\mathbf{x}_{i}^{ans})$

\Else
    \State Construct the tool-use context $\mathbf{x}_{i}^{tool}$ using the global image $I_i^{G}$ and task query $q_i^G$
    \State Set the ground-truth tool-use response target $\mathbf{y}_{i}^{tool}=\texttt{\textless think\textgreater}\rightarrow\texttt{\textless tool\_call\textgreater}$
    \State Provide the ground-truth cropped image $I_i^{C}$ and the inspection query $q_i^{C}$ as the second visual input
    \State Construct the answer context $\mathbf{x}_{i}^{ans}$ using $I_i^{G}$, $q_i^G$, $\mathbf{y}_{i}^{tool}$, $I_i^{C}$, and $q_i^{C}$
    \State Set the ground-truth final response target $\mathbf{y}_{i}^{ans}=\texttt{\textless think\textgreater}\rightarrow\texttt{\textless answer\textgreater}$
    \State Form the supervised sequence $\mathbf{s}_i=[\mathbf{x}_{i}^{tool},\mathbf{y}_{i}^{tool},\mathbf{x}_{i}^{ans},\mathbf{y}_{i}^{ans}]$
    \State Unmask only the assistant tokens in $\mathbf{y}_{i}^{tool}$ and $\mathbf{y}_{i}^{ans}$
    \State Set $\mathcal{L}_{total}\leftarrow \mathcal{L}(\mathbf{y}_{i}^{tool}\mid\mathbf{x}_{i}^{tool})+\mathcal{L}(\mathbf{y}_{i}^{ans}\mid\mathbf{x}_{i}^{ans})$
\EndIf

\State Mask all system, user, image, and padding tokens in $\mathbf{s}_i$ from the loss
\State Compute the masked assistant-token loss $\mathcal{L}_{total}$
\State Update only the LoRA parameters $\phi$ using $\nabla_{\phi}\mathcal{L}_{total}$

\EndFor
\end{algorithmic}
\end{algorithm}

Given an image $I_i^{G}$ and a task query $q_i^G$, we obtain the downsampled global image $\widetilde{I}_i^{G}$ using a downsampling factor $s$.
The first-turn context is then constructed using $\widetilde{I}_i^{G}$ and $q_i^G$. Since many regions in wide-view construction-site images are irrelevant to a specific query, using $\widetilde{I}_i^{G}$ reduces unnecessary visual tokens from redundant background regions while still providing the overall scene context needed for global reasoning. The model then generates the first-turn response:

\vspace{-1em}
\begin{equation}
\hat{\mathbf{y}}_{i}^{(1)}
\sim
p_{\theta_0,\phi}
\left(
\cdot
\mid
\widetilde{I}_i^{G}, q_i^G
\right)    
\end{equation}
\vspace{-1em}

The role of this first-turn response depends on the generated tag. If $\hat{\mathbf{y}}_{i}^{(1)}$ contains an \texttt{\textless answer\textgreater} tag, we treat it as the final answer response $\hat{\mathbf{y}}_{i}^{ans}$. The model then directly returns the final task output without additional visual inspection. This direct-answer path is used when the downsampled global image provides sufficient evidence for the given query.

If $\hat{\mathbf{y}}_{i}^{(1)}$ contains a \texttt{\textless tool\_call\textgreater} tag, we treat it as the model-generated tool-use response $\hat{\mathbf{y}}_{i}^{tool}$. AVA-VLM then parses the predicted crop box from $\hat{\mathbf{y}}_{i}^{tool}$. Let the predicted box in the downsampled image coordinate system be $\hat{b}_{i}^{C} =[\hat{x}_{1},\hat{y}_{1},\hat{x}_{2},\hat{y}_{2}]$. Since the first-turn image is downsampled, the predicted crop box is mapped back to the original image coordinate system before cropping as
$\hat{b}_{i}^{G} = s \cdot \hat{b}_{i}^{C}.$
The mapped box $\hat{b}_{i}^{G}$ is then used to crop a local region $\hat{I}_{i}^{C}$ from the original full-resolution image $I_i^{G}$.
This step allows AVA-VLM to recover high-resolution local evidence only for the region that the model identifies as relevant to the query. After obtaining the predicted crop, AVA-VLM constructs the second-turn context using the preceding model-generated tool-use response $\hat{\mathbf{y}}_{i}^{tool}$, the cropped image $\hat{I}_{i}^{C}$, and the crop-inspection query $q_i^{C}$. The model then generates the final crop-conditioned response:

\begin{equation}
\hat{\mathbf{y}}_{i}^{ans}
\sim
p_{\theta_0,\phi}
\left(
\cdot
\mid
\widetilde{I}_i^{G}, q_i^G, \hat{\mathbf{y}}_{i}^{tool}, \hat{I}_{i}^{C}, q_i^{C}
\right)
\end{equation}

\vspace{-1.5em}
\section{Experiment}

We conduct our evaluation on ConstructionSite10K~\cite{ConstructionSite10K}, as it is the only benchmark that includes both mid- and long-distance observations, which are essential for evaluating the operational-range challenges considered in this study.

\subsection{Implementation Details}
\label{subsec:ImplementationDetails}

\paragraph{\textbf{Training setup.}}
All construction-tailored VLMs are trained using LoRA tuning on the training and validation splits curated in Section~\ref{sec:Dataset}. All tuned models use the same pretrained Qwen2.5-VL backbone and the same curated dataset. All models are trained on two NVIDIA RTX A6000 GPUs with 48GB memory. We use the same training hyperparameters across tuned models: weight decay of 0.01, warmup ratio of 0.03, LoRA rank of 128, LoRA alpha of 256, LoRA dropout of 0.05, a maximum of 10 training epochs, and early stopping with a patience of 3. For all methods, the final response is trained to be enclosed by \texttt{\textless answer\textgreater} and \texttt{\textless /answer\textgreater}, and this tag is used to extract the final output during evaluation.

\paragraph{\textbf{Compared methods.}}
We compare AVA-VLM with three Qwen2.5-VL (7B)-based settings as described below. The purpose of this comparison is not to compare different backbone architectures, but to evaluate how different adaptation strategies behave under the same backbone, dataset, and tuning configuration. Except for the Qwen2.5-VL-based settings described below, the results of the remaining zero-shot and few-shot methods are reported from \cite{ConstructionSite10K}.

\begin{itemize}
\item \textbf{Qwen2.5}: the pretrained general-purpose Qwen2.5-VL model without construction-site adaptation. The model receives the global image and task query and returns the final answer as $I_i^{G}, q_i^G \rightarrow \texttt{\textless answer\textgreater}.$ This setting evaluates the zero-shot transfer ability of a general-purpose VLM on construction-site monitoring tasks.

\item \textbf{Qwen2.5 (Baseline)}: the main construction-site-tailored baseline. It follows the direct-QA adaptation strategy widely adopted in existing construction-site VLM studies, where the model directly maps the global image and task query to the final answer as $I_i^{G}, q_i^G \rightarrow \texttt{\textless answer\textgreater}.$ This setting evaluates the performance of the most commonly used construction-site-specific VLM tuning strategy.

\item \textbf{Qwen2.5$^{\star}$}: a few-shot example image-based CoT baseline. It provides an example image and its reasoning trace as additional visual and textual context before answering the target query as $I^{E}, r_i^{E}, I_i^{G}, q_i^G \rightarrow \texttt{\textless think\textgreater} \rightarrow \texttt{\textless answer\textgreater},$ where $r_i^{E}$ denotes the reasoning trace for the example image $I_i^{E}$. This setting is included to examine whether example image-based reasoning can further improve performance, while also quantifying the additional visual-token cost introduced by the example image.

\item \textbf{AVA-VLM (Ours)}: Our AVA-VLM adapts the same Qwen2.5-VL backbone using the proposed training framework, where the model learns to decide whether the global visual evidence is sufficient or whether a query-relevant local crop should be inspected. This setting evaluates whether adaptive crop-based reasoning improves robustness across operation ranges, scale robustness, and inference efficiency under the same backbone.
\end{itemize}

\paragraph{\textbf{Evaluation metrics.}}
Following the benchmark~\cite{ConstructionSite10K}, we evaluate AVA-VLM and the compared methods on three construction-site monitoring tasks: violation identification (VI), object detection (OD), and image captioning (IC).

For the VI task, we evaluate classification performance using precision, recall, and F1-score. We further evaluate the quality of the model responses for samples identified as violations. For this, we use an LLM judge~\cite{llama} to measure the semantic alignment between the model-generated response and the ground-truth explanation, assigning a score from 0 to 6, where a higher score indicates stronger semantic agreement. For localization quality, we compute the Intersection over Union (IoU) between the predicted violation bounding box and the corresponding ground-truth bounding box.

For the OD task, we evaluate localization performance using micro-averaged IoU. IoU measures the spatial overlap between the predicted and ground-truth bounding boxes. For each object category, we report two evaluation settings: Object Exist and Total. Object Exist IoU is computed only on samples where the target object exists in the ground truth, and therefore measures how accurately a model localizes objects once localization is required. Total IoU is computed over all samples, including both object-present and object-absent cases, and therefore evaluates practical object-detection behavior by considering not only localization accuracy but also whether the model correctly handles queries for absent objects. We additionally report Avg. IoU by averaging the six micro-averaged IoU values across the three object categories and the two evaluation settings.

For the IC task, we evaluate description quality using three complementary captioning and semantic similarity metrics: SPICE~\cite{SPICE}, METEOR~\cite{METEOR}, and BERTScore~\cite{BertScore}. SPICE measures the semantic content of generated captions by comparing scene-graph-level propositions, making it suitable for evaluating whether the generated description captures important objects, attributes, and relations in the construction scene. METEOR evaluates lexical and phrase-level alignment with the reference caption while accounting for flexible word matching, and therefore captures whether the generated description is linguistically consistent with the human annotation. BERTScore measures contextual semantic similarity between the generated and reference captions, providing a more flexible evaluation of paraphrased but semantically equivalent descriptions. Please refer to the ConstructionSite10K benchmark~\cite{ConstructionSite10K} for more details.

\subsection{Violation Identification Performance Analysis}
\label{subsec:VI}

\begin{table*}[t]
\centering
\caption{Per-class precision, recall, F1-score, and visual-token comparison on the VI task. AVA-VLM uses 1/4 downsampled global images while other models use full-size global images.}
\label{tab:VI_class_overall}
\resizebox{\textwidth}{!}{%
\begin{tabular}{lcccccccc}
\toprule
\multirow{2}{*}{Model} 
& \multicolumn{3}{c}{No Violation $\uparrow$} 
& \multicolumn{3}{c}{PPE Violation $\uparrow$} 
& \multirow{2}{*}{Overall F1. $\uparrow$}
& \multirow{2}{*}{Visual Tokens $\downarrow$} \\
\cmidrule(lr){2-4}
\cmidrule(lr){5-7}
& Precision & Recall & F1-score
& Precision & Recall & F1-score
&  &  \\
\midrule

\multicolumn{9}{c}{Zero-shot / Few-shot} \\
GPT~\cite{GPT} 
& 97.0 & 24.6 & 39.2 
& 20.4 & 76.4 & 32.2 
& 35.7
& -- \\

GPT 5-shot~\cite{GPT} 
& 98.9 & 32.0 & 48.4 
& 18.2 & 89.4 & 30.2 
& 39.3
& -- \\

LLaVA-v1.5 (13B)~\cite{LLaVA} 
& 88.0 & 43.4 & 58.1 
& 12.0 & 54.0 & 19.6 
& 38.9
& -- \\

LLaVA-v1.5 (13B) CoT~\cite{LLaVA-CoT} 
& 91.0 & 44.0 & 59.3 
& 12.0 & 55.0 & 19.7 
& 39.5
& -- \\

LLaVA-NeXT (34B) 1-shot~\cite{LLaVA-NeXT} 
& 87.1 & 88.7 & 87.9 
& 14.3 & 13.0 & 13.6 
& 50.8
& -- \\

Qwen2.5 (7B)~\cite{Qwen2_5}        
& 89.5 & 100 & 94.5 
& 0 & 0 & 0 
& 47.3
& 100\% \\

\midrule
\multicolumn{9}{c}{Construction-tailored VLMs (LoRA-tuning)} \\
Qwen2.5 (Baseline) (7B)~\cite{Qwen2_5}         
& 91.1 & \underline{\textbf{99.9}} & 95.3 
& \underline{\textbf{94.4}} & 16.8 & 28.6 
& 62.0 
& 100\%  \\

Qwen2.5$^\star$ (7B)~\cite{Qwen2_5} 
& \underline{\textbf{94.3}} & 92.4 & 93.3 
& 44.5 & \underline{\textbf{52.1}} & 48.0 
& 70.7 \textcolor{gray}{(+8.7$\uparrow$)}
& 193\% \textcolor{gray}{(+93\%$\uparrow$)}\\

AVA-VLM (7B) (Ours)            
& 93.8 & 97.6 & \underline{\textbf{95.7}} 
& 68.5 & 45.2 & \underline{\textbf{54.5}} 
& \underline{\textbf{75.1}} \textcolor{gray}{(+13.1$\uparrow$)}
& \underline{\textbf{30.6\%}} \textcolor{gray}{(-69.4\%$\downarrow$)} \\

\midrule
Human upper bound 
& 95.3 & 98.3 & 96.8 
& 95.6 & 66.6 & 78.5 
& 88.0
& -- \\
\bottomrule
\end{tabular}%
}
\end{table*}

\subsubsection{Violation Classification Performance}

\paragraph{\textbf{Overall Performance.}}
Table~\ref{tab:VI_class_overall} compares the overall violation-identification performance and visual-token usage across zero-shot, few-shot, and construction-tailored VLMs. The zero-shot Qwen2.5 model fails to identify PPE violations, as shown by its no-violation recall of 100.0\% and PPE-violation precision, recall, and F1-score of 0.0\%. This indicates that the general-purpose VLM predicts all samples as no violation, resulting in an overall F1-score of only 47.3\%. After LoRA tuning with direct QA supervision, Qwen2.5 (Baseline) improves the overall F1-score to 62.0\%, confirming the importance of construction-specific adaptation. 

Moreover, providing an example image with a reasoning-and-answer trace further improves model performance. Qwen2.5$^\star$ increases the overall F1-score from 62.0\% to 70.7\%, corresponding to an 8.7 percentage-point improvement over the direct QA baseline. It also improves PPE-violation recall from 16.8\% to 52.1\%, showing that example image-based CoT can help the model identify more violation cases. However, this improvement is accompanied by substantial visual-token overhead, increasing visual-token usage from 100\% to 193\%. 

In contrast, AVA-VLM achieves the best overall performance while substantially reducing visual-token consumption. Compared with the direct QA baseline, AVA-VLM improves the overall F1-score by 13.1 percentage points, from 62.0\% to 75.1\%, and achieves the highest PPE-violation F1-score of 54.5\%. At the same time, AVA-VLM reduces visual-token usage by 69.4\%, using only 30.6\% of the baseline visual-token budget. This efficiency gain is explained by the selective tool-use behavior of AVA-VLM. As shown in Figure~\ref{fig:VQA_tool_call}(a), AVA-VLM directly answers 82.0\% of VI samples without calling the crop tool, indicating that the downsampled global image is sufficient for most cases. For the remaining 18.0\% of samples, the model requests a local crop and performs detailed inspection only on the query-relevant region. Therefore, AVA-VLM does not need to process every image at full resolution. Instead, it uses a 1/4 downsampled global image, which accounts for 25.0\% of the baseline visual tokens, and adds only 5.6\% additional visual tokens from cropped images. These results demonstrate that AVA-VLM improves performance by selectively allocating high-resolution visual processing to samples where the global evidence is insufficient.

\begin{table*}[t]
\centering
\caption{Camera-distance-wise per-class precision, recall, F1-score, and visual-token comparison on the VI task. Visual-token usage is normalized by Qwen2.5 (Baseline) within each distance group.}
\label{tab:VI_class_distance}
\resizebox{\textwidth}{!}{%
\begin{tabular}{lcccccccc}
\toprule
\multirow{2}{*}{\textbf{Model}}
& \multicolumn{3}{c}{\textbf{No Violation $\uparrow$}}
& \multicolumn{3}{c}{\textbf{PPE Violation $\uparrow$}}
& \multirow{2}{*}{\textbf{Overall F1. $\uparrow$}}
& \multirow{2}{*}{\textbf{Visual Tokens $\downarrow$}} \\
\cmidrule(lr){2-4}
\cmidrule(lr){5-7}
& \textbf{Precision} & \textbf{Recall} & \textbf{F1-score}
& \textbf{Precision} & \textbf{Recall} & \textbf{F1-score}
&  &  \\
\midrule

\midrule
\multicolumn{9}{c}{\textbf{Short Distance}} \\
Qwen2.5 (Baseline) (7B)~\cite{Qwen2_5}
& 90.7 & \underline{\textbf{99.9}} & 95.1
& \underline{\textbf{96.8}} & 20.0 & 33.1
& 64.1
& \underline{\textbf{100\%}}  \\

Qwen2.5$^\star$ (7B)~\cite{Qwen2_5}
& 94.5 & 91.6 & 93.0
& 46.8 & 58.0 & 51.8
& 72.4 \textcolor{gray}{(+8.3$\uparrow$)}
& 187\% \textcolor{gray}{(+87\%$\uparrow$)} \\

AVA-VLM (7B) (Ours)
& \underline{\textbf{95.6}} & 96.9 & \underline{\textbf{96.3}}
& 73.1 & \underline{\textbf{65.3}} & \underline{\textbf{69.0}}
& \underline{\textbf{82.7}} \textcolor{gray}{(+18.6$\uparrow$)}
& 103\% \textcolor{gray}{(+3\%$\uparrow$)} \\

\midrule
\multicolumn{9}{c}{\textbf{Mid Distance}} \\
Qwen2.5 (Baseline) (7B)~\cite{Qwen2_5}
& 90.0 & \underline{\textbf{99.8}} & 95.2
& \underline{\textbf{90.5}} & 14.6 & 25.2
& 60.2
& \underline{\textbf{100\%}} \\

Qwen2.5$^\star$ (7B)~\cite{Qwen2_5}
& 94.0 & 93.5 & 93.7
& 46.3 & 48.5 & 47.4
& 70.6 \textcolor{gray}{(+10.4$\uparrow$)}
& 196\% \textcolor{gray}{(+96\%$\uparrow$)} \\

AVA-VLM (7B) (Ours)
& \underline{\textbf{95.9}} & 97.5 & \underline{\textbf{96.7}}
& 75.0 & \underline{\textbf{64.6}} & \underline{\textbf{69.4}}
& \underline{\textbf{83.1}} \textcolor{gray}{(+22.9$\uparrow$)}
& 109\% \textcolor{gray}{(+9\%$\uparrow$)} \\

\midrule
\multicolumn{9}{c}{\textbf{Long Distance}} \\
Qwen2.5 (Baseline) (7B)~\cite{Qwen2_5}
& 93.4 & \underline{\textbf{100}} & 96.6
& \underline{\textbf{100}} & 8.7 & 16.0
& 56.3
& \underline{\textbf{100\%}} \\

Qwen2.5$^\star$ (7B)~\cite{Qwen2_5}
& 94.8 & 91.6 & 93.2
& 24.2 & 34.8 & 28.6
& 60.9 \textcolor{gray}{(+4.6$\uparrow$)}
& 218\% \textcolor{gray}{(+118\%$\uparrow$)} \\

AVA-VLM (7B) (Ours)
& \underline{\textbf{97.0}} & 97.7 & \underline{\textbf{97.3}}
& 66.7 & \underline{\textbf{60.9}} & \underline{\textbf{63.6}}
& \underline{\textbf{80.5}} \textcolor{gray}{(+24.2$\uparrow$)}
& 109\% \textcolor{gray}{(+9\%$\uparrow$)} \\

\bottomrule
\end{tabular}%
}
\end{table*}

\begin{center}
\includegraphics[
    width=\textwidth,
    height=\textheight,
    keepaspectratio
]{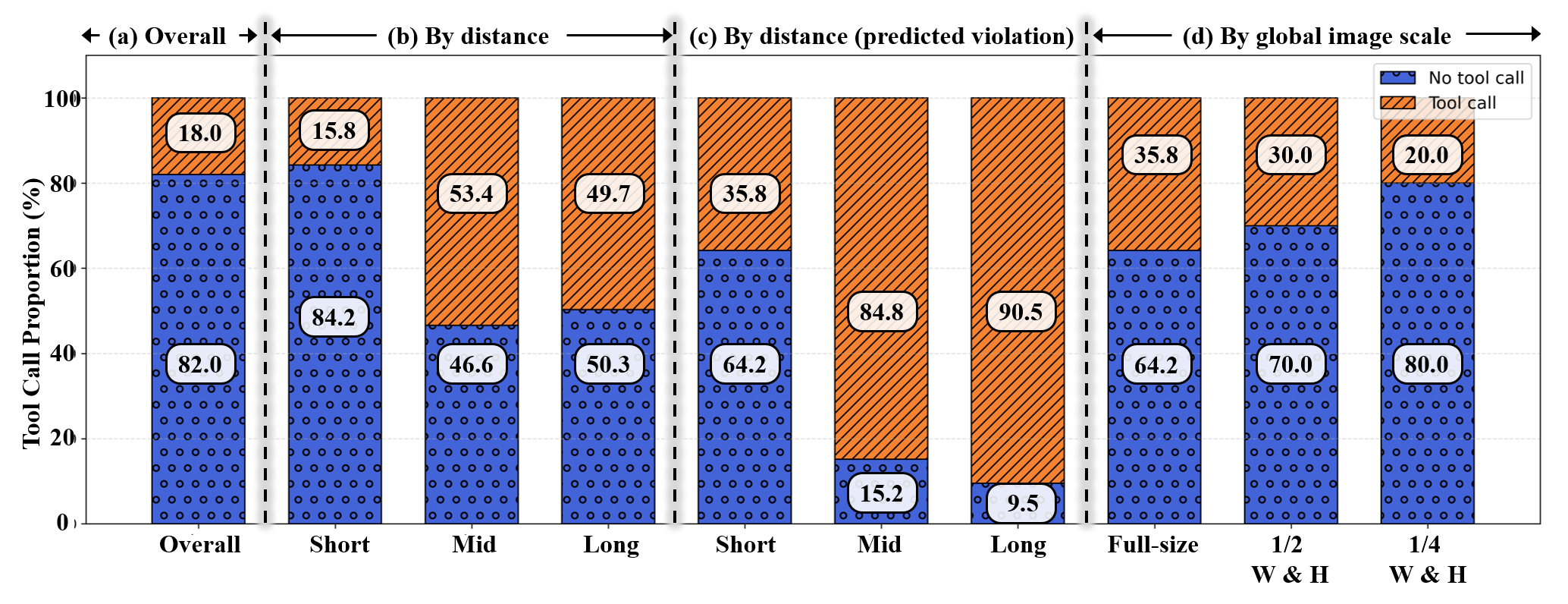}
\vspace{-0.8cm}
\captionof{figure}{
Tool-call usage ratio of AVA-VLM on the VI task, showing selective local inspection across overall samples, camera distances, predicted PPE-violation samples, and reduced global image resolutions.
}
\label{fig:VQA_tool_call}
\end{center}

\paragraph{\textbf{Camera-Distance-wise Performance.}} Table~\ref{tab:VI_class_distance} further analyzes VI performance across camera distances. Qwen2.5 (Baseline) shows clear performance degradation as the camera distance increases, with the overall F1-score decreasing from 64.1\% at short distance to 56.3\% at long distance. This degradation is mainly caused by the poor recall for PPE violations. This indicates that direct QA-based adaptation tends to classify visually ambiguous or small distant workers as no violation, limiting its operational range in wide-view monitoring scenarios.

Providing an example image with a reasoning-and-answer trace partially mitigates this limitation. Qwen2.5$^\star$ improves the overall F1-score over the direct QA baseline at all distances, with gains of 8.3, 10.4, and 4.6 percentage points for short-, mid-, and long-distance images, respectively. It also improves the PPE-violation F1-score compared with the baseline, especially at short and mid distances. However, this improvement requires a large increase in visual-token usage, reaching 187\%, 196\%, and 218\% of the baseline at short-, mid-, and long-distance, respectively. This suggests that example image-based CoT can provide useful visual and reasoning context, but it does not fully resolve the difficulty of recognizing small or distant violation evidence and incurs substantial additional visual-token cost.

In contrast, AVA-VLM achieves consistently strong performance across all camera distances. It obtains the highest overall F1-score for short, mid, and long distances, achieving 82.7\%, 83.1\%, and 80.5\%, respectively. Compared with the direct QA baseline, AVA-VLM improves the overall F1-score by 18.6, 22.9, and 24.2 percentage points for short-, mid-, and long-distance images, respectively. The improvement is particularly notable at long distance, where AVA-VLM increases the PPE-violation F1-score from 16.0\% to 63.6\%. This demonstrates that the proposed adaptive visual attention mechanism effectively extends the operational range of construction-site VLMs by enabling detailed inspection of small or distant violation evidence.

The strong distance-wise performance of AVA-VLM is achieved with only a small increase in visual-token usage. As shown in Table~\ref{tab:VI_class_distance}, AVA-VLM uses only 3\%, 9\%, and 9\% additional visual tokens compared with the direct QA baseline for short-, mid-, and long-distance images, respectively. This efficiency is explained by the selective cropping behavior shown in Figure~\ref{fig:VQA_tool_call}(b). AVA-VLM rarely calls the crop tool for short-distance samples, where the relevant evidence is usually sufficiently visible from the global image, but it uses the crop tool more frequently for mid- and long-distance samples, where workers and PPE details are more likely to be small or ambiguous. Furthermore, Figure~\ref{fig:VQA_tool_call}(c) shows that tool-call usage becomes much more frequent for samples predicted as PPE violations, especially at mid and long distances. This indicates that AVA-VLM selectively allocates local high-resolution inspection to safety-critical cases where detailed visual evidence is needed, allowing it to maintain strong PPE-violation identification performance across camera distances without substantially increasing the visual-token budget.

\begin{table*}[t]
\centering
\caption{Global-image-scale-wise per-class precision, recall, F1-score, and visual-token comparison on the VI task. Visual-token usage is normalized by Qwen2.5 (Baseline) using full-size global images.}
\label{tab:VI_class_scale}
\resizebox{\textwidth}{!}{%
\begin{tabular}{lcccccccc}
\toprule
\multirow{2}{*}{\textbf{Model}} 
& \multicolumn{3}{c}{\textbf{No Violation $\uparrow$}} 
& \multicolumn{3}{c}{\textbf{PPE Violation $\uparrow$}} 
& \multirow{2}{*}{\textbf{Overall F1. $\uparrow$}}
& \multirow{2}{*}{\textbf{Visual Tokens $\downarrow$}} \\
\cmidrule(lr){2-4}
\cmidrule(lr){5-7}
& \textbf{Precision} & \textbf{Recall} & \textbf{F1-score}
& \textbf{Precision} & \textbf{Recall} & \textbf{F1-score}
&  &  \\
\midrule

\midrule
\multicolumn{9}{c}{\textbf{Full-Size Global Image}} \\
Qwen2.5 (Baseline) (7B)~\cite{Qwen2_5}         
& 91.1 & \underline{\textbf{99.9}} & 95.3 
& \underline{\textbf{94.4}} & 16.8 & 28.6 
& 62.0 
& \underline{\textbf{100\%}}  \\

Qwen2.5$^\star$ (7B)~\cite{Qwen2_5} 
& 94.3 & 92.4 & 93.3 
& 44.5 & 52.1 & 48.0 
& 70.7 \textcolor{gray}{(+8.7$\uparrow$)}
& 193\% \textcolor{gray}{(+93\%$\uparrow$)}\\

AVA-VLM (7B) (Ours)            
& \underline{\textbf{95.9}} & 97.3 & \underline{\textbf{96.6}} 
& 73.4 & \underline{\textbf{64.7}} & \underline{\textbf{68.8}} 
& \underline{\textbf{82.7}} \textcolor{gray}{(+20.7$\uparrow$)}
& 106\% \textcolor{gray}{(+6\%$\uparrow$)} \\

\midrule
\multicolumn{9}{c}{\textbf{Global Image Downsampled to 1/2 Width and Height}} \\
Qwen2.5 (Baseline) (7B)~\cite{Qwen2_5}          
& 90.4 & \underline{\textbf{99.8}} & 94.9
& \underline{\textbf{87.1}} & 0.09 & 16.2
& 55.6
& \underline{\textbf{25\%}}\\

Qwen2.5$^\star$ (7B)~\cite{Qwen2_5}    
& 93.6 & 93.7 & 93.6 
& 45.5 & 44.9 & 45.2 
& 69.4 \textcolor{gray}{(+13.8$\uparrow$)}
& 47.4\% \textcolor{gray}{(+22.4\%$\uparrow$)} \\

AVA-VLM (7B) (Ours)            
& \underline{\textbf{93.8}} & 97.6 & \underline{\textbf{95.7}} 
& 68.5 & \underline{\textbf{45.2}} & \underline{\textbf{54.5}} 
& \underline{\textbf{75.1}} \textcolor{gray}{(+19.5$\uparrow$)}
& 30.6\% \textcolor{gray}{(+5.6\%$\uparrow$)} \\

\midrule
\multicolumn{9}{c}{\textbf{Global Image Downsampled to 1/4 Width and Height}} \\
Qwen2.5 (Baseline) (7B)~\cite{Qwen2_5}          
& 90.0 & \underline{\textbf{99.9}} & \underline{\textbf{94.7}} 
& \underline{\textbf{83.3}} & 5.0 & 9.3 
& 52.0
& \underline{\textbf{6.4\%}} \\

Qwen2.5$^\star$ (7B)~\cite{Qwen2_5}     
& 91.3 & 90.8 & 91.1 
& 25.1 & 26.4 & 25.7 
& 58.4 \textcolor{gray}{(+6.4$\uparrow$)}
& 12.4\% \textcolor{gray}{(+6.0\%$\uparrow$)} \\

AVA-VLM (7B) (Ours)            
& \underline{\textbf{92.0}} & 96.1 & 94.0 
& 45.7 & \underline{\textbf{28.4}} & \underline{\textbf{35.0}} 
& \underline{\textbf{64.5}} \textcolor{gray}{(+12.5$\uparrow$)}
& 12.7\% \textcolor{gray}{(+6.2\%$\uparrow$)}\\

\bottomrule
\end{tabular}%
}
\end{table*}

\paragraph{\textbf{Performance under Reduced Global Image Resolution.}}
Table~\ref{tab:VI_class_scale} evaluates VI performance under different global image resolutions by downsampling the width and height of the global image. As the global image resolution decreases, all methods show performance degradation, indicating that reduced-resolution inputs make PPE-violation evidence harder to recognize. For Qwen2.5 (Baseline), the overall F1-score decreases from 62.0\% with full-size global images to 52.0\% when the global image width and height are downsampled to 1/4. This degradation is mainly caused by the reduction in PPE-violation performance, where the PPE-violation F1-score decreases from 28.6\% to 9.3\%. This result shows that direct QA-based adaptation becomes increasingly biased toward predicting no violation when safety-critical visual evidence is degraded by reduced input resolution.

Qwen2.5$^\star$ also shows a similar trend. Although example image-based CoT improves performance compared with the direct QA baseline at each global image resolution, its overall F1-score decreases from 70.7\% with full-size global images to 58.4\% with 1/4-width-and-height global images. This indicates that providing an example image with a reasoning-and-answer trace can partially mitigate the reduced-resolution problem, but it cannot fully recover visual details that are lost in the target image. Moreover, Qwen2.5$^\star$ still requires additional visual tokens from the example image, which limits its efficiency advantage under reduced-resolution settings.

In contrast, AVA-VLM consistently outperforms the other methods across all global image resolutions. Even when the global image width and height are downsampled to 1/4, AVA-VLM achieves an overall F1-score of 64.5\%, which is higher than Qwen2.5 (Baseline) using full-size global images. This demonstrates that AVA-VLM provides more reliable VI performance under reduced global image resolution by combining low-resolution global reasoning with adaptive local inspection. Nevertheless, AVA-VLM also shows performance degradation as the global image resolution decreases, with the overall F1-score decreasing from 82.7\% with full-size global images to 75.1\% with 1/2-width-and-height global images and 64.5\% with 1/4-width-and-height global images. This degradation suggests that overly aggressive global-image downsampling can make the first-turn global observation insufficient for deciding whether detailed inspection is needed, causing the model to directly predict no violation rather than triggering the crop tool. 

This trend is also reflected in Figure~\ref{fig:VQA_tool_call}(d). As the global image width and height are reduced, the tool-call ratio decreases from 35.8\% with full-size global images to 30.0\% with 1/2-width-and-height global images and 20.0\% with 1/4-width-and-height global images. This suggests that, under very low-resolution global inputs, some violation-relevant regions become too visually ambiguous to trigger the crop tool, leading the model to directly answer from the global image and increasing missed PPE violations. Based on this trade-off, we use the 1/2-width-and-height global image setting with adaptive local inspection as the default AVA-VLM configuration in the overall comparison reported in Table~\ref{tab:VI_class_overall}. The qualitative results in Figure~\ref{fig:VQA_main_example} further support this finding. AVA-VLM either directly answers from the downsampled global image when the evidence is sufficiently visible (Figure~\ref{fig:VQA_main_example} Top) or requests a local crop when detailed inspection is needed (Figure~\ref{fig:VQA_main_example} Bottom), enabling reliable PPE-violation identification with fewer visual tokens.

\begin{center}
\includegraphics[
    width=\textwidth,
    height=\textheight,
    keepaspectratio
]{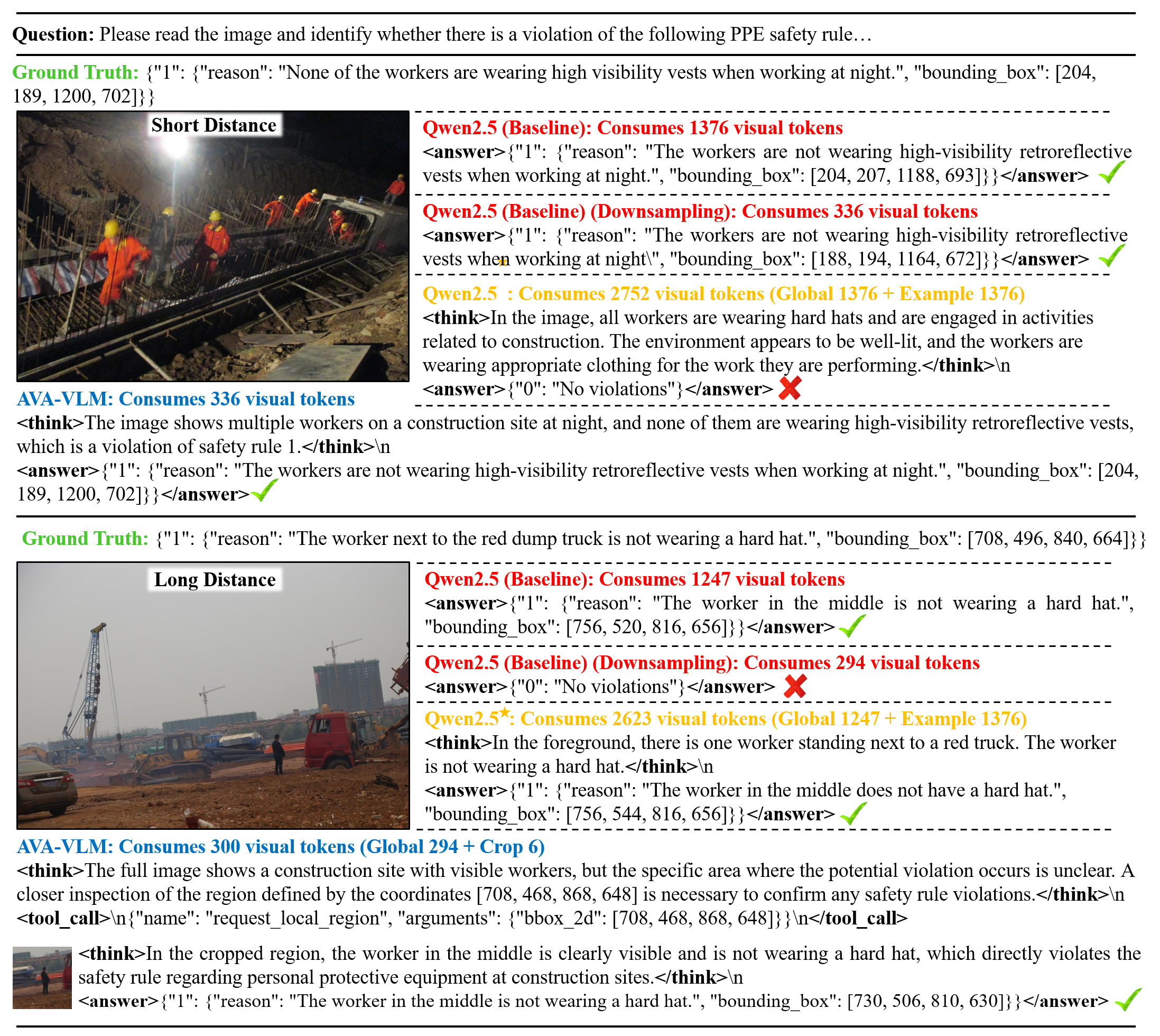}
\vspace{-0.5cm}
\captionof{figure}{
Qualitative results of PPE-violation identification, showing how AVA-VLM selectively uses local inspection when global evidence is insufficient.
}
\label{fig:VQA_main_example}
\end{center}

\subsubsection{Violation Reasoning and Localization Performance}

\paragraph{\textbf{Overall Performance.}}
Table~\ref{tab:VI_IoU_overall} evaluates the reasoning and localization quality for PPE-violation identification. The reasoning quality refers to how well the generated textual response explains the PPE-violation evidence in a manner consistent with the ground-truth annotation. We quantify this using the Average Mark, which is an LLM-judge score ranging from 0 to 6; a higher score indicates stronger semantic agreement between the model-generated explanation and the ground-truth violation rationale. On the other hand, localization quality refers to how accurately the model identifies the spatial region associated with the violation. We measure this using IoU-based metrics: true positive (TP) IoU evaluates localization accuracy only for correctly predicted PPE-violation cases, Obj. Exist IoU evaluates localization performance over ground-truth violation samples where a violation object exists, thereby also reflecting missed violations, and Total IoU further reflects overall localization performance across the full evaluation set.

Table~\ref{tab:VI_IoU_overall} shows that the zero-shot Qwen2.5 model does not predict any PPE-violation samples, resulting in a correct prediction ratio of 0\% and zero localization performance. This result is consistent with the classification results in Table~\ref{tab:VI_class_overall}, where the zero-shot model predicts all samples as no violation. After construction-specific LoRA tuning, the models produce substantially better reasoning quality. This indicates that, once a PPE violation is correctly identified, the tuned models can generate high-quality textual explanations for the predicted violation.

However, reasoning quality alone does not guarantee reliable violation identification and localization. Qwen2.5 (Baseline) achieves the highest TP IoU of 68.7\%, suggesting that it localizes violations well when it correctly predicts a PPE violation. Nevertheless, its correct prediction ratio is only 16.8\%, indicating that the model identifies PPE violations only in highly confident cases and misses many violations. As a result, its Obj. Exist IoU and Total IoU remain low at 24.1\% and 23.3\%, respectively. This shows that direct QA-based adaptation can produce accurate localization for easy positive cases, but it lacks sufficient recall over all PPE-violation samples.

\begin{table*}[t]
\centering
\caption{Overall qualities of reasoning and localization for model-predicted PPE violations on the VI task. TP IoU is computed on correctly predicted PPE-violation samples, Obj. Exist IoU is computed on all ground-truth PPE-violation samples, and Total IoU includes negative samples. AVA-VLM uses 1/4 downsampled global images while other models use full-size global images.}
\label{tab:VI_IoU_overall}
\resizebox{\textwidth}{!}{
\begin{tabular}{lcccccc}
\toprule
\textbf{Model} 
& \textbf{Correct Pred. Ratio $\uparrow$} 
& \textbf{Average Mark $\uparrow$} 
& \textbf{TP IoU $\uparrow$} 
& \textbf{Obj. Exist IoU $\uparrow$}
& \textbf{Total IoU $\uparrow$}
& \textbf{Visual Tokens $\downarrow$} \\
\midrule

\multicolumn{7}{c}{\textbf{Zero-shot / Few-shot}} \\
GPT~\cite{GPT}                       
& -- & 4.5 & 11.9 & -- & -- & -- \\
GPT 5-shot~\cite{GPT}                
& -- & 4.7 & 14.0 & -- & -- & -- \\
LLaVA-v1.5 (13B)~\cite{LLaVA}        
& -- & 2.9 & 20.0 & -- & -- & -- \\
LLaVA-v1.5 (13B) CoT~\cite{LLaVA-CoT} 
& -- & 3.1 & 6.9 & -- & -- & -- \\
LLaVA-NeXT (34B) 1-shot~\cite{LLaVA-NeXT} 
& -- & 3.9 & -- & -- & -- & -- \\
Qwen2.5 (7B)~\cite{Qwen2_5}    
& 0 & 0 & 0 & 0 & 0 & 100\% \\

\midrule
\multicolumn{7}{c}{\textbf{Construction-tailored VLMs (LoRA-tuning)}} \\

Qwen2.5 (Baseline) (7B)~\cite{Qwen2_5}    
& 16.8  
& \underline{\textbf{5.6}} 
& \underline{\textbf{68.7}} 
& 24.1 
& 23.3 
& 100\% \\

Qwen2.5$^{\star}$ (7B)~\cite{Qwen2_5}    
& \underline{\textbf{52.1}} \textcolor{gray}{(+35.3$\uparrow$)} 
& 5.1 \textcolor{gray}{(-0.5$\downarrow$)} 
& 48.8 \textcolor{gray}{(-19.9$\downarrow$)} 
& \underline{\textbf{35.2}} \textcolor{gray}{(+11.1$\uparrow$)} 
& 24.1 \textcolor{gray}{(+0.8$\uparrow$)} 
& 193\% \textcolor{gray}{(+93\%$\uparrow$)} \\

AVA-VLM (7B) (Ours)                         
& 45.2 \textcolor{gray}{(+28.4$\uparrow$)} 
& 5.4 \textcolor{gray}{(-0.2$\downarrow$)} 
& 50.9 \textcolor{gray}{(-17.8$\downarrow$)} 
& 33.9 \textcolor{gray}{(+9.8$\uparrow$)} 
& \underline{\textbf{28.4}} \textcolor{gray}{(+5.1$\uparrow$)} 
& \underline{\textbf{30.6\%}} \textcolor{gray}{(-69.4\%$\downarrow$)} \\

\bottomrule
\end{tabular}%
}
\end{table*}

Qwen2.5$^{\star}$ reduces missed violations by increasing the correct prediction ratio from 16.8\% to 52.1\%. This leads to a higher Obj. Exist IoU of 35.2\%, showing that example image-based CoT helps the model identify more ground-truth PPE violations. However, this improvement comes with a decrease in localization accuracy for correctly predicted violations, as TP IoU drops from 68.7\% to 48.8\%. In addition, Qwen2.5$^{\star}$ requires 193\% visual tokens, nearly doubling the visual-token budget compared with the direct QA baseline. Therefore, example image-based CoT improves violation coverage but sacrifices localization quality and inference efficiency.

In contrast, AVA-VLM provides a more balanced trade-off between violation identification, localization, and visual-token efficiency. AVA-VLM achieves a correct prediction ratio of 45.2\%, substantially higher than Qwen2.5 (Baseline), while maintaining a TP IoU of 50.9\%. Although its TP IoU is lower than that of Qwen2.5 (Baseline), AVA-VLM improves Obj. Exist IoU from 24.1\% to 33.9\% and achieves the highest Total IoU of 28.4\%. This indicates that AVA-VLM not only detects more PPE-violation cases than the direct QA baseline, but also maintains reasonable localization quality across both positive and negative samples. Importantly, AVA-VLM achieves this balanced reasoning and localization performance using only 30.6\% of the baseline visual-token budget. These results demonstrate that AVA-VLM improves PPE-violation reasoning and localization by selectively using local high-resolution inspection when the global evidence is insufficient.

\paragraph{\textbf{Camera-Distance-wise Performance.}}
Table~\ref{tab:VI_IoU_distance} analyzes the reasoning and localization quality across camera distances. As the camera distance increases, Qwen2.5 (Baseline) shows substantial degradation in both violation identification coverage and localization quality. Its correct prediction ratio decreases from 20.0\% at short distance to 8.7\% at long distance. Although the baseline achieves high TP IoU for short- and mid-distance samples, its TP IoU drops sharply to 36.1\% at long distance. More importantly, its Obj. Exist IoU and Total IoU decrease to only 5.9\% at long distance, indicating that the model not only misses most long-range PPE violations but also fails to localize the responsible workers reliably when violations are detected.

Qwen2.5$^{\star}$ partially improves violation coverage by using an example image and reasoning-and-answer trace. Compared with Qwen2.5 (Baseline), it increases the correct prediction ratio across all camera distances, reaching 58.0\%, 48.5\%, and 34.8\% for short-, mid-, and long-distance samples, respectively. However, this improvement does not consistently translate into better localization quality. For example, at short and mid distances, Qwen2.5$^{\star}$ improves Obj. Exist IoU but reduces TP IoU compared with the baseline, suggesting that it detects more violation cases but localizes the responsible workers less accurately among correctly predicted samples. At long distance, its Total IoU remains only 4.3\%, despite using 218\% visual tokens. This shows that example image-based CoT increases violation coverage but still struggles with long-range localization and incurs substantial visual-token overhead.

In contrast, AVA-VLM maintains strong reasoning and localization performance across camera distances. It achieves the highest correct prediction ratio at all distances, with 65.3\%, 64.6\%, and 60.9\% for short-, mid-, and long-distance samples, respectively. The advantage of AVA-VLM is particularly clear at long distance, where it improves the correct prediction ratio by 52.2\% over the direct QA baseline. It also achieves the highest long-distance TP IoU, Obj. Exist IoU, and Total IoU, reaching 48.1\%, 46.4\%, and 42.6\%, respectively. These results indicate that AVA-VLM can not only identify more long-range PPE violations but also better localize the responsible workers by using adaptive local inspection, demonstrating that AVA-VLM provides a more effective balance between violation coverage, localization quality, and inference efficiency in challenging wide-view monitoring scenarios.

\begin{table*}[t]
\centering
\caption{Camera-distance-wise reasoning and localization quality for model-predicted PPE violations on the VI task. Visual-token usage is normalized by Qwen2.5 (Baseline) within each distance group.}
\label{tab:VI_IoU_distance}
\resizebox{\textwidth}{!}{%
\begin{tabular}{lcccccc}
\toprule
\textbf{Model} 
& \textbf{Correct Pred. Ratio$\uparrow$} 
& \textbf{Average Mark $\uparrow$} 
& \textbf{TP IoU $\uparrow$}
& \textbf{Obj. Exist IoU $\uparrow$}
& \textbf{Total IoU $\uparrow$}
& \textbf{Visual Tokens $\downarrow$} \\
\midrule

\midrule
\multicolumn{7}{c}{\textbf{Short Distance}} \\

Qwen2.5 (Baseline) (7B)~\cite{Qwen2_5}    
& 20.0 
& \underline{\textbf{5.5}} 
& \underline{\textbf{69.1}} 
& 24.5 
& 23.9 
& \underline{\textbf{100\%}} \\

Qwen2.5$^{\star}$ (7B)~\cite{Qwen2_5}    
& 58.0 \textcolor{gray}{(+38.0$\uparrow$)} 
& 5.1 \textcolor{gray}{(-0.4$\downarrow$)} 
& 52.0 \textcolor{gray}{(-17.1$\downarrow$)} 
& 37.5 \textcolor{gray}{(+13.0$\uparrow$)} 
& 25.5 \textcolor{gray}{(+1.6$\uparrow$)} 
& 187\% \textcolor{gray}{(+87\%$\uparrow$)} \\

AVA-VLM (7B) (Ours)                         
& \underline{\textbf{65.3}} \textcolor{gray}{(+45.3$\uparrow$)} 
& \underline{\textbf{5.5}} \textcolor{gray}{(+0.0)} 
& 52.0 \textcolor{gray}{(-17.1$\downarrow$)} 
& \underline{\textbf{41.2}} \textcolor{gray}{(+16.7$\uparrow$)} 
& \underline{\textbf{35.6}} \textcolor{gray}{(+11.7$\uparrow$)} 
& 103\% \textcolor{gray}{(+3\%$\uparrow$)} \\

\midrule
\multicolumn{7}{c}{\textbf{Mid Distance}} \\

Qwen2.5 (Baseline) (7B)~\cite{Qwen2_5}    
& 14.6 
& \underline{\textbf{5.8}} 
& \underline{\textbf{71.4}} 
& 24.4 
& 22.9 
& \underline{\textbf{100\%}} \\

Qwen2.5$^{\star}$ (7B)~\cite{Qwen2_5}    
& 48.5 \textcolor{gray}{(+33.9$\uparrow$)} 
& 5.3 \textcolor{gray}{(-0.5$\downarrow$)} 
& 45.8 \textcolor{gray}{(-25.6$\downarrow$)} 
& 30.7 \textcolor{gray}{(+6.3$\uparrow$)} 
& 22.8 \textcolor{gray}{(-0.1$\downarrow$)} 
& 196\% \textcolor{gray}{(+96\%$\uparrow$)} \\

AVA-VLM (7B) (Ours)                         
& \underline{\textbf{64.6}} \textcolor{gray}{(+50.0$\uparrow$)} 
& 5.5 \textcolor{gray}{(-0.3$\downarrow$)} 
& 45.5 \textcolor{gray}{(-25.9$\downarrow$)} 
& \underline{\textbf{33.1}} \textcolor{gray}{(+8.7$\uparrow$)} 
& \underline{\textbf{30.4}} \textcolor{gray}{(+7.5$\uparrow$)} 
& 109\% \textcolor{gray}{(+9\%$\uparrow$)} \\

\midrule
\multicolumn{7}{c}{\textbf{Long Distance}} \\

Qwen2.5 (Baseline) (7B)~\cite{Qwen2_5}    
& 8.7 
& 5.0 
& 36.1 
& 5.9 
& 5.9
& \underline{\textbf{100\%}} \\

Qwen2.5$^{\star}$ (7B)~\cite{Qwen2_5}    
& 34.8 \textcolor{gray}{(+26.1$\uparrow$)} 
& 5.0 \textcolor{gray}{(+0.0)} 
& 38.0 \textcolor{gray}{(+1.9$\uparrow$)} 
& 12.4 \textcolor{gray}{(+6.5$\uparrow$)} 
& 4.3 \textcolor{gray}{(-1.6$\downarrow$)} 
& 218\% \textcolor{gray}{(+118\%$\uparrow$)} \\

AVA-VLM (7B) (Ours)                         
& \underline{\textbf{60.9}} \textcolor{gray}{(+52.2$\uparrow$)} 
& \underline{\textbf{5.4}} \textcolor{gray}{(+0.4$\uparrow$)} 
& \underline{\textbf{48.1}} \textcolor{gray}{(+12.0$\uparrow$)} 
& \underline{\textbf{46.4}} \textcolor{gray}{(+40.5$\uparrow$)} 
& \underline{\textbf{42.6}} \textcolor{gray}{(+36.7$\uparrow$)} 
& 109\% \textcolor{gray}{(+9\%$\uparrow$)} \\

\bottomrule
\end{tabular}%
}
\end{table*}

\paragraph{\textbf{Performance under Reduced Global Image Resolution.}}
Table~\ref{tab:VI_IoU_scale} evaluates the reasoning and localization performance under reduced global image resolutions. As the global image width and height are downsampled, Qwen2.5 (Baseline) shows a sharp degradation in violation identification coverage. Its correct prediction ratio decreases from 16.8\% to 4.9\% with 1/4-width-and-height global images. Although the baseline maintains high TP IoU among the few correctly predicted violation samples, its Obj. Exist IoU and Total IoU decrease substantially under downsampling. In particular, when the global image width and height are downsampled to 1/4, the baseline achieves only 7.1\% Obj. Exist IoU and 6.9\% Total IoU. This indicates that direct QA-based adaptation becomes highly unreliable under reduced-resolution inputs because it misses most PPE violations before localization can be evaluated.

Qwen2.5$^{\star}$ partially mitigates this degradation by using an example image and reasoning-and-answer trace. It achieves higher correct prediction ratios than the baseline at all image scales, reaching 52.1\%, 44.9\%, and 26.4\% for full-size, 1/2-width-and-height, and 1/4-width-and-height global images, respectively. However, its localization quality becomes less stable as the global image resolution decreases. The TP IoU drops from 48.8\% at full size to 17.1\% at 1/4 width and height. Moreover, Qwen2.5$^{\star}$ requires additional visual tokens from the example image, using 193\%, 47.4\%, and 12.4\% of the full-size baseline visual-token budget across the three settings. These results show that example image-based CoT improves violation coverage under downsampling, but it does not reliably preserve localization quality and still incurs additional visual-token cost.

In contrast, AVA-VLM maintains more robust localization performance across reduced global image resolutions. With full-size global images, AVA-VLM achieves the highest correct prediction ratio, Obj. Exist IoU, and Total IoU, reaching 64.7\%, 39.3\%, and 34.5\%, respectively. More importantly, AVA-VLM remains effective under downsampling. When the global image width and height are downsampled to 1/2, AVA-VLM achieves a correct prediction ratio of 45.2\%, an Obj. Exist IoU of 33.9\%, and the highest Total IoU of 28.4\%, while using only 30.6\% of the full-size baseline visual-token budget. Even under the more aggressive 1/4-width-and-height setting, AVA-VLM achieves the best correct prediction ratio, Exist IoU, and Total IoU among all methods, with 28.3\%, 26.6\%, and 21.6\%, respectively.

These results demonstrate that AVA-VLM is more robust to reduced global image resolution than direct QA-based adaptation and example image-based CoT. Although AVA-VLM also experiences performance degradation as the global image becomes severely downsampled, it consistently preserves a better balance between identifying PPE violations and localizing responsible workers.

The qualitative results in Figure~\ref{fig:VQA_downsampling_example} further illustrate this robustness under global image downsampling. In the full-size image, all methods can identify the PPE violation because the responsible workers remain sufficiently visible. However, as the global image width and height are downsampled, Qwen2.5 (Baseline) fails to identify the violation, showing that direct QA-based inference becomes unreliable when the violation evidence is visually degraded. Qwen2.5$^{\star}$ can sometimes preserve violation identification by using the example image at the downsampled resolution, but its localization becomes unstable under stronger downsampling. In contrast, AVA-VLM consistently requests a local crop around the ambiguous worker region and recovers high-resolution evidence from the original image. This enables AVA-VLM to correctly identify the PPE violation and localize the responsible workers even when the downsampled global image alone is insufficient, supporting the quantitative improvements in the correct prediction ratio, Obj. TP IoU, Exist IoU and Total IoU shown in Table~\ref{tab:VI_IoU_scale}.

\begin{table*}[t]
\centering
\caption{Global-image-scale-wise reasoning and localization for model-predicted PPE violations on the VI task. Visual-token usage is normalized by Qwen2.5 (Baseline) using full-size global images.}
\label{tab:VI_IoU_scale}
\resizebox{\textwidth}{!}{%
\begin{tabular}{lcccccc}
\toprule
\textbf{Model} 
& \textbf{Correct Pred. Ratio$\uparrow$} 
& \textbf{Average Mark $\uparrow$} 
& \textbf{TP IoU $\uparrow$}
& \textbf{Obj. Exist IoU $\uparrow$}
& \textbf{Total IoU $\uparrow$}
& \textbf{Visual Tokens $\downarrow$} \\
\midrule

\midrule
\multicolumn{7}{c}{\textbf{Full-Size Global Image}} \\

Qwen2.5 (Baseline) (7B)~\cite{Qwen2_5}    
& 16.8 
& \underline{\textbf{5.6}} 
& \underline{\textbf{68.7}} 
& 24.1 
& 23.3 
& \underline{\textbf{100\%}} \\

Qwen2.5$^{\star}$ (7B)~\cite{Qwen2_5}    
& 52.1 \textcolor{gray}{(+35.3$\uparrow$)}  
& 5.1 \textcolor{gray}{(-0.5$\downarrow$)} 
& 48.8 \textcolor{gray}{(-19.9$\downarrow$)} 
& 35.2 \textcolor{gray}{(+11.1$\uparrow$)} 
& 24.1 \textcolor{gray}{(+0.8$\uparrow$)} 
& 193\% \textcolor{gray}{(+93\%$\uparrow$)} \\

AVA-VLM (7B) (Ours)                         
& \underline{\textbf{64.7}} \textcolor{gray}{(+47.9$\uparrow$)} 
& 5.4 \textcolor{gray}{(-0.2$\downarrow$)} 
& 49.0 \textcolor{gray}{(-19.7$\downarrow$)} 
& \underline{\textbf{39.3}} \textcolor{gray}{(+15.2$\uparrow$)} 
& \underline{\textbf{34.5}} \textcolor{gray}{(+11.2$\uparrow$)} 
& 106\% \textcolor{gray}{(+6\%$\uparrow$)} \\

\midrule
\multicolumn{7}{c}{\textbf{Global Image Downsampled to 1/2 Width and Height}} \\

Qwen2.5 (Baseline) (7B)~\cite{Qwen2_5}    
& 8.9
& \underline{\textbf{5.7}}
& \underline{\textbf{73.2}}
& 16.7 
& 16.1
& \underline{\textbf{25\%}} \\

Qwen2.5$^{\star}$ (7B)~\cite{Qwen2_5}    
& 44.9 \textcolor{gray}{(+36.0$\uparrow$)} 
& 5.0 \textcolor{gray}{(-0.7$\downarrow$)} 
& 41.6 \textcolor{gray}{(-31.6$\downarrow$)} 
& 28.2 \textcolor{gray}{(+11.5$\uparrow$)}
& 18.5 \textcolor{gray}{(+2.4$\uparrow$)}
& 47.4\% \textcolor{gray}{(+22.4\%$\uparrow$)} \\

AVA-VLM (7B) (Ours)                         
& \underline{\textbf{45.2}} \textcolor{gray}{(+36.3$\uparrow$)} 
& 5.5 \textcolor{gray}{(-0.2$\downarrow$)} 
& 50.9 \textcolor{gray}{(-22.3$\downarrow$)} 
& \underline{\textbf{33.9}} \textcolor{gray}{(+17.2$\uparrow$)}
& \underline{\textbf{28.4}} \textcolor{gray}{(+12.3$\uparrow$)}
& 30.6\% \textcolor{gray}{(+5.6\%$\uparrow$)} \\

\midrule
\multicolumn{7}{c}{\textbf{Global Image Downsampled to 1/4 Width and Height}} \\

Qwen2.5 (Baseline) (7B)~\cite{Qwen2_5}    
& 4.9
& \underline{\textbf{5.8}}
& \underline{\textbf{61.0}}
& 7.1 
& 6.9 
& \underline{\textbf{6.4\%}} \\

Qwen2.5$^{\star}$ (7B)~\cite{Qwen2_5}    
& 26.4 \textcolor{gray}{(+21.5$\uparrow$)} 
& 4.8 \textcolor{gray}{(-1.0$\downarrow$)} 
& 17.1 \textcolor{gray}{(-43.9$\downarrow$)} 
& 13.6 \textcolor{gray}{(+6.5$\uparrow$)}
& 8.4  \textcolor{gray}{(+1.5$\uparrow$)}
& 12.4\% \textcolor{gray}{(+6.0\%$\uparrow$)} \\

AVA-VLM (7B) (Ours)                         
& \underline{\textbf{28.3}} \textcolor{gray}{(+23.4$\uparrow$)} 
& 5.3 \textcolor{gray}{(-0.5$\downarrow$)} 
& 53.5 \textcolor{gray}{(-7.5$\downarrow$)} 
& \underline{\textbf{26.6}} \textcolor{gray}{(+19.5$\uparrow$)}
& \underline{\textbf{21.6}} \textcolor{gray}{(+14.7$\uparrow$)}
& 12.7\% \textcolor{gray}{(+6.3\%$\uparrow$)} \\

\bottomrule
\end{tabular}%
}
\end{table*}

\begin{center}
\includegraphics[
    width=\textwidth,
    height=\textheight,
    keepaspectratio
]{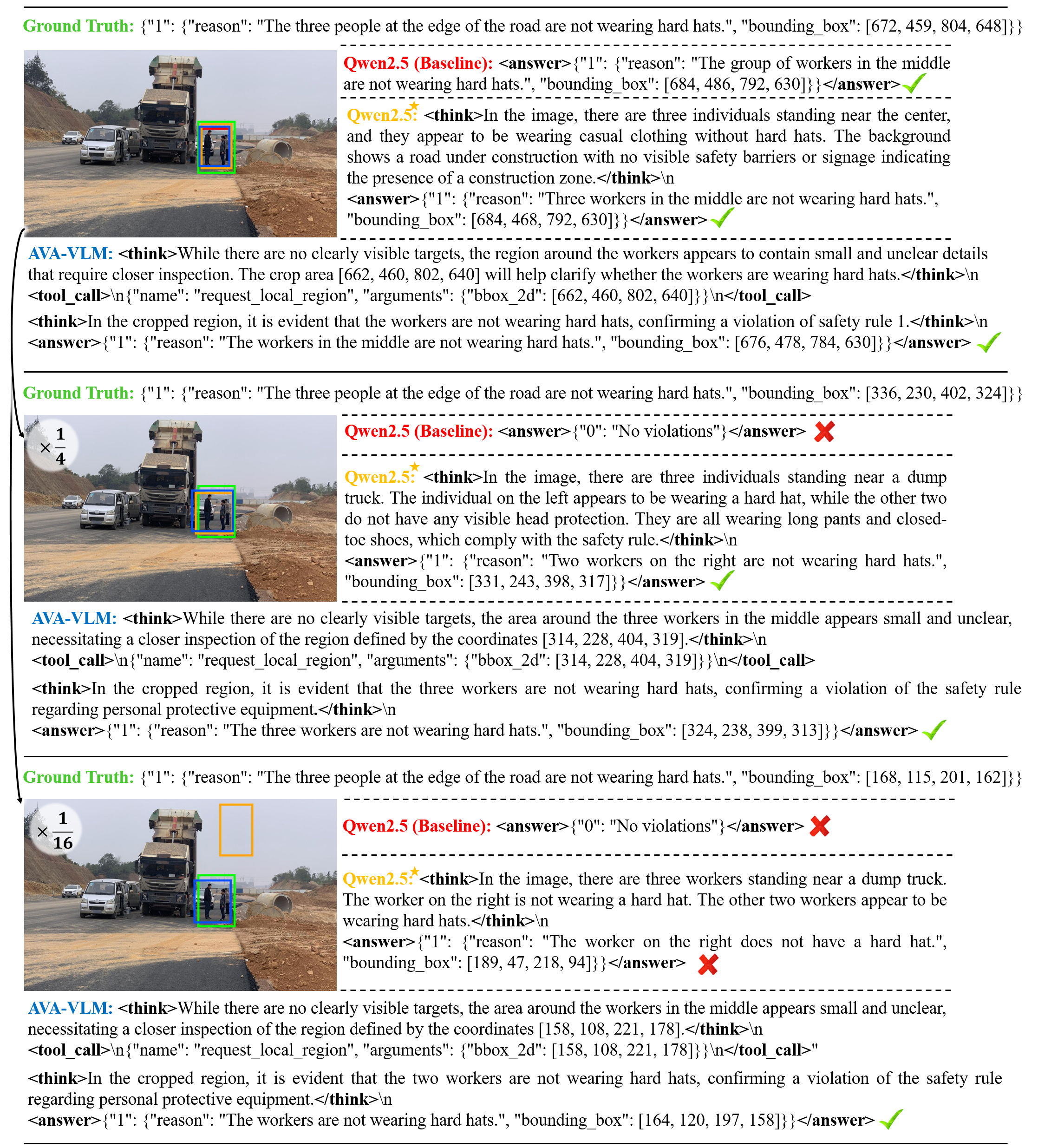}

\captionof{figure}{
Qualitative results of PPE-violation identification under global image downsampling, showing AVA-VLM's robustness when global evidence becomes insufficient.
}
\label{fig:VQA_downsampling_example}
\end{center}

\begin{table*}[t]
\centering
\caption{Micro-averaged OD IoU comparison across model types. Object Exist denotes IoU computed only on samples where the target object exists in the ground truth, while Total denotes IoU computed with negative samples included. Avg. IoU is computed by averaging the six micro-averaged IoU values across object categories and evaluation settings. AVA-VLM uses 1/4 downsampled global images while other models use full-size global images.}
\label{tab:constructionsite_object_detection}
\resizebox{0.99\textwidth}{!}{%
\begin{tabular}{lcccccccc}
\toprule
\multirow{2}{*}{\textbf{Model}} 
& \multicolumn{2}{c}{\textbf{Excavators $\uparrow$}} 
& \multicolumn{2}{c}{\textbf{Rebars $\uparrow$}} 
& \multicolumn{2}{c}{\textbf{Workers w/ white hard hats $\uparrow$}}
& \multirow{2}{*}{\textbf{Avg. IoU $\uparrow$}}
& \multirow{2}{*}{\textbf{Visual Tokens $\downarrow$}} \\
\cmidrule(lr){2-3}
\cmidrule(lr){4-5}
\cmidrule(lr){6-7}
& \textbf{Object Exist} & \textbf{Total}
& \textbf{Object Exist} & \textbf{Total}
& \textbf{Object Exist} & \textbf{Total}
&  &  \\
\midrule
\midrule
\multicolumn{9}{c}{\textbf{Zero-shot / Few-shot}} \\
GPT~\cite{GPT}
& 35.8 & 29.3 & 18.2 & 11.6 & 10.1 & 8.6 & 18.9 & -- \\

LLaVA-v1.5 (7B)~\cite{LLaVA}
& 34.6 & 20.9 & 13.3 & 3.1 & 15.0 & 3.8 & 15.1 & -- \\

LLaVA-v1.5 (13B)~\cite{LLaVA}
& 54.5 & 33.9 & 19.0 & 3.8 & 23.9 & 4.2 & 23.2 & -- \\

Grounding DINO~\cite{GroundingDino}
& 71.0 & 45.4 & 15.3 & 2.0 & 11.2 & 2.1 & 24.5 & -- \\

Qwen2.5 (7B)~\cite{Qwen2_5}
& 83.7 & 65.1 & 24.1 & 5.6 & 53.7 & 22.8 & 42.5 & 100\% \\

\midrule
\multicolumn{9}{c}{\textbf{Construction-tailored VLMs (LoRA-tuning)}} \\
Qwen2.5 (Baseline) (7B)~\cite{Qwen2_5}
& \underline{\textbf{87.0}} & \underline{\textbf{86.8}} 
& \underline{\textbf{47.3}} & \underline{\textbf{46.0}} 
& \underline{\textbf{72.1}} & \underline{\textbf{69.3}} 
& \underline{\textbf{68.1}} 
& 100\%  \\

Qwen2.5$^{\star}$ (7B)~\cite{Qwen2_5}
& 84.7 & 79.0 
& 36.7 & 17.6 
& 60.9 & 59.3 
& 56.4 \textcolor{gray}{(-11.7$\downarrow$)}
& 193\% \textcolor{gray}{(+93\%$\uparrow$)} \\

AVA-VLM (7B) (Ours)
& 79.5 & 79.4 
& 38.3 & 37.4 
& 64.6 & 62.6 
& 60.3 \textcolor{gray}{(-7.8$\downarrow$)}
& \underline{\textbf{25.8\%}} \textcolor{gray}{(-74.2\%$\downarrow$)} \\

\bottomrule
\end{tabular}%
}
\end{table*}

\begin{center}
\includegraphics[
    width=\textwidth,
    height=\textheight,
    keepaspectratio
]{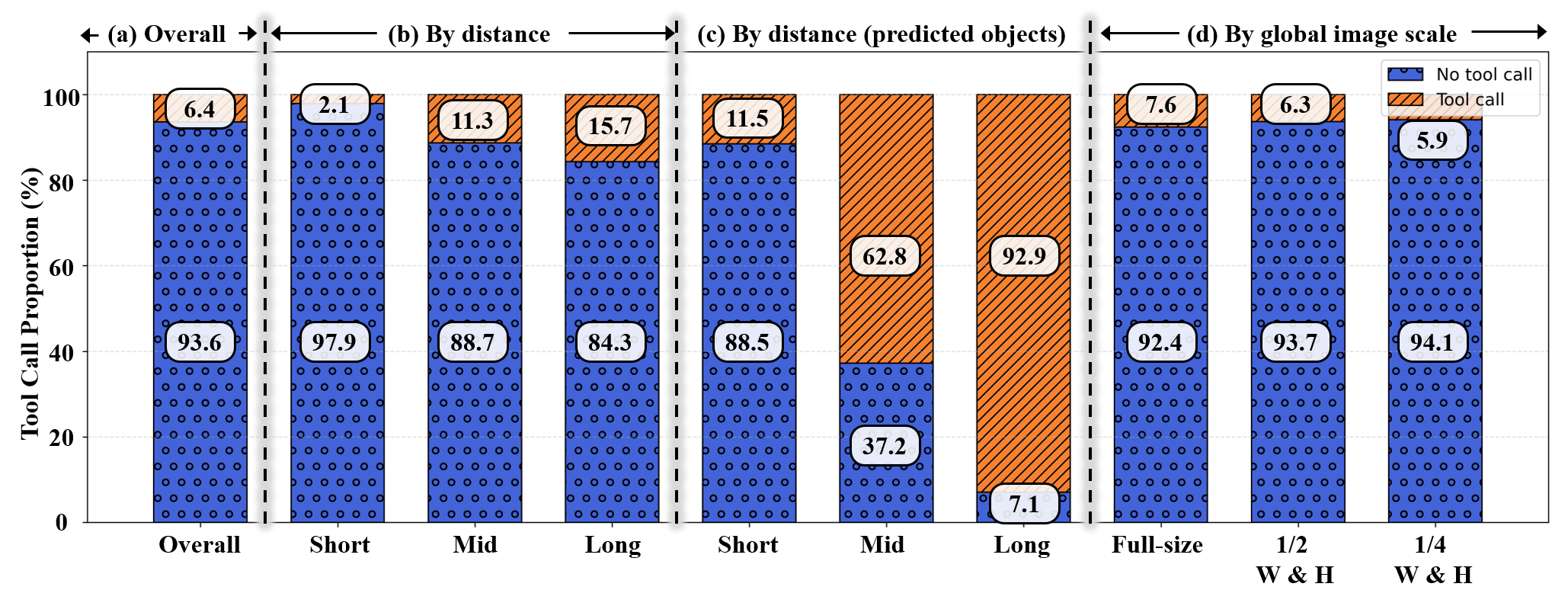}

\captionof{figure}{
Tool-call usage ratio of AVA-VLM on the OD task, showing selective local inspection across overall samples, camera distances, predicted object-present samples, and reduced global image resolutions.
}
\label{fig:OD_tool_call_ratio}
\end{center}

\subsection{Object Detection Performance Analysis}
\label{subsec:OD}

\paragraph{\textbf{Overall Performance.}}
Table~\ref{tab:constructionsite_object_detection} compares the overall object detection performance and visual-token usage across zero-shot, few-shot, and construction-tailored models. Among the zero-shot and few-shot models, Qwen2.5 achieves the strongest performance, with an average IoU of 42.5\%, outperforming GPT, LLaVA variants, and Grounding DINO. However, after construction-specific LoRA tuning, Qwen2.5 (Baseline) substantially improves the average IoU to 68.1\%, achieving the best Object Exist and Total IoU across all three object categories. This confirms that direct construction-site adaptation is effective for the OD task when full-size global images are used.

In contrast to the VI task, providing an example image with a reasoning-and-answer trace does not improve OD performance. Qwen2.5$^{\star}$ decreases the average IoU from 68.1\% to 56.4\%, while increasing visual-token usage from 100\% to 193\%. This degradation suggests that example image-based CoT can make the OD task more difficult rather than easier. The additional example image and reasoning context may distract the model from the target-image localization task, leading to lower IoU despite the much larger visual-token budget.

AVA-VLM also shows lower overall OD performance than the full-resolution Qwen2.5 (Baseline), with an average IoU of 60.3\% compared with 68.1\%. However, AVA-VLM achieves this performance using only 25.8\% of the baseline visual-token budget, corresponding to a 74.2\% reduction in visual-token usage. Compared with Qwen2.5$^{\star}$, AVA-VLM achieves a higher average IoU while using substantially fewer visual tokens. This indicates that AVA-VLM provides a more favorable performance--efficiency trade-off than example image-based CoT for OD.

The performance gap between AVA-VLM and the full-resolution direct QA baseline suggests that the OD task is highly dependent on high-resolution visual information. In particular, when Qwen2.5 (Baseline) uses 1/4 downsampled global images, its average IoU decreases to 60.0\%, which is close to the 60.3\% achieved by AVA-VLM (see Table~\ref{tab:OD_scale}). This suggests that the lower overall OD performance of AVA-VLM is largely associated with reduced global-image resolution and limited local inspection, rather than a failure of construction-specific adaptation.

This behavior is also reflected in the tool-use pattern of AVA-VLM. As shown in Figure~\ref{fig:OD_tool_call_ratio} (a), AVA-VLM calls the crop tool for only 6.4\% of OD samples overall, meaning that most predictions are made from the downsampled global image without local high-resolution inspection. This tool-call ratio is much lower than that observed in the VI task, suggesting that AVA-VLM may under-utilize local inspection for OD. Therefore, while AVA-VLM substantially reduces visual-token usage, its overall OD performance reflects a trade-off between efficiency and localization accuracy. In the following analyses, we further examine whether this trade-off becomes more favorable under challenging camera-distance and reduced-resolution settings, where adaptive local inspection is expected to provide greater benefit.

\paragraph{\textbf{Camera-Distance-wise Performance.}}

\begin{table*}[t]
\centering
\caption{Camera-distance-wise micro-averaged OD IoU comparison across object categories. Object Exist denotes IoU computed only on samples where the target object exists, Total includes negative samples, and visual-token usage is normalized by Qwen2.5 (Baseline) within each distance group.}
\label{tab:OD_distance}
\resizebox{0.99\textwidth}{!}{%
\begin{tabular}{lcccccccc}
\toprule
\multirow{2}{*}{\textbf{Model}} 
& \multicolumn{2}{c}{\textbf{Excavators $\uparrow$}} 
& \multicolumn{2}{c}{\textbf{Rebars $\uparrow$}} 
& \multicolumn{2}{c}{\textbf{Workers w/ white hard hats $\uparrow$}}
& \multirow{2}{*}{\textbf{Avg. IoU $\uparrow$}}
& \multirow{2}{*}{\textbf{Visual Tokens $\downarrow$}} \\
\cmidrule(lr){2-3}
\cmidrule(lr){4-5}
\cmidrule(lr){6-7}
& \textbf{Object Exist} & \textbf{Total}
& \textbf{Object Exist} & \textbf{Total}
& \textbf{Object Exist} & \textbf{Total}
&  &  \\
\midrule

\midrule
\multicolumn{9}{c}{\textbf{Short Distance}} \\
Qwen2.5 (Baseline) (7B)~\cite{Qwen2_5}
& \underline{\textbf{88.2}} & \underline{\textbf{88.0}} 
& \underline{\textbf{47.8}} & \underline{\textbf{47.1}} 
& 73.1 & 70.4 
& \underline{\textbf{69.1}} 
& \underline{\textbf{100\%}} \\

Qwen2.5$^{\star}$ (7B)~\cite{Qwen2_5}
& 86.4 & 82.0 
& 41.2 & 19.6 
& 64.1 & 62.4 
& 59.3 \textcolor{gray}{(-9.8$\downarrow$)}
& 187\% \textcolor{gray}{(+87\%$\uparrow$)} \\

AVA-VLM (7B) (Ours)
& 87.5 & 87.0 
& 43.2 & 40.9 
& \underline{\textbf{74.4}} & \underline{\textbf{71.1}} 
& 67.4 \textcolor{gray}{(-1.7$\downarrow$)}
& \underline{\textbf{100\%}} \textcolor{gray}{(+0\%)} \\

\midrule
\multicolumn{9}{c}{\textbf{Mid Distance}} \\
Qwen2.5 (Baseline) (7B)~\cite{Qwen2_5}
& \underline{\textbf{84.9}} & \underline{\textbf{84.7}} 
& 50.0 & \underline{\textbf{48.6}} 
& 66.5 & 63.3
& \underline{\textbf{66.3}} 
& \underline{\textbf{100\%}} \\

Qwen2.5$^{\star}$ (7B)~\cite{Qwen2_5}
& 83.1 & 74.7 
& 35.5 & 16.5 
& 40.9 & 39.5 
& 48.4 \textcolor{gray}{(-17.9$\downarrow$)}
& 196\% \textcolor{gray}{(+96\%$\uparrow$)} \\

AVA-VLM (7B) (Ours)
& 79.1 & 79.1 
& \underline{\textbf{51.6}} & 48.0 
& \underline{\textbf{67.9}} & \underline{\textbf{64.9}} 
& 65.1 \textcolor{gray}{(-1.2$\downarrow$)}
& 103\% \textcolor{gray}{(+3\%$\uparrow$)} \\

\midrule
\multicolumn{9}{c}{\textbf{Long Distance}} \\
Qwen2.5 (Baseline) (7B)~\cite{Qwen2_5}
& \underline{\textbf{72.0}} & \underline{\textbf{71.7}} 
& 38.6 & 37.1 
& 40.7 & 39.8 
& 50.0 
& \underline{\textbf{100\%}} \\

Qwen2.5$^{\star}$ (7B)~\cite{Qwen2_5}
& 53.9 & 45.7 
& 33.5 & 17.8 
& 4.2 & 4.2 
& 26.6 \textcolor{gray}{(-23.4$\downarrow$)}
& 218\% \textcolor{gray}{(+118\%$\uparrow$)} \\

AVA-VLM (7B) (Ours)
& 70.3 & 70.1 
& \underline{\textbf{46.3}} & \underline{\textbf{43.6}} 
& \underline{\textbf{42.4}} & \underline{\textbf{41.4}} 
& \underline{\textbf{52.4}} \textcolor{gray}{(+2.4$\uparrow$)}
& 106\% \textcolor{gray}{(+6\%$\uparrow$)} \\

\bottomrule
\end{tabular}%
}
\end{table*}

Table~\ref{tab:OD_distance} analyzes OD performance across camera distances. Overall, the full-resolution Qwen2.5 (Baseline) maintains strong performance at short and mid distances, achieving average IoUs of 69.1\% and 66.3\%, respectively. However, its performance drops substantially at long distance, where the average IoU decreases to 50.0\%. This degradation indicates that long-range object detection remains challenging even for a construction-tailored VLM when objects become smaller and less visually distinct in the global image.

Qwen2.5$^{\star}$ shows a more severe performance drop as camera distance increases. Its average IoU decreases from 59.3\% at short distance to 26.6\% at long distance, despite using substantially more visual tokens than the direct QA baseline. This result suggests that example image-based CoT does not provide a reliable benefit for OD. In particular, the additional example image and reasoning trace may distract the model from precise target-image localization, causing larger degradation for small or distant objects.

In contrast, at short and mid distances, AVA-VLM achieves average IoUs of 67.4\% and 65.1\%, which are close to the full-resolution Qwen2.5 (Baseline) while using comparable visual-token budgets. More importantly, AVA-VLM achieves the best long-distance performance, improving the average IoU from 50.0\% to 52.4\% over the baseline with only 6\% additional visual tokens. This suggests that adaptive local inspection becomes more beneficial in long-range OD scenarios, where target objects are smaller and global-image evidence is less sufficient. This trend is also reflected in Figure~\ref{fig:OD_tool_call_ratio}(b). AVA-VLM calls the crop tool more frequently as the camera distance increases, with the tool-call ratio rising from 2.1\% at short distance to 11.3\% at mid distance and 15.7\% at long distance. Moreover, Figure~\ref{fig:OD_tool_call_ratio}(c) shows that, among samples where AVA-VLM predicts object presence, the tool-call ratio increases much more sharply with distance, reaching 62.8\% at mid distance and 92.9\% at long distance. This indicates that local high-resolution inspection is especially useful when the model needs to predict bounding boxes for distant objects. Therefore, although AVA-VLM uses the crop tool conservatively overall, it increasingly relies on local inspection in challenging long-range cases where detailed visual evidence is needed for localization.

Object-wise, AVA-VLM is particularly effective for smaller object categories. For workers with white hard hats, AVA-VLM achieves the best Object Exist and Total IoU at all camera distances, outperforming Qwen2.5 (Baseline) at short, mid, and long distances. This is important because workers are typically smaller and more difficult to localize than large equipment such as excavators. AVA-VLM also improves long-distance rebar detection, achieving the highest Object Exist and Total IoU for rebars at long distance. These results show that adaptive visual attention is especially beneficial for small-object and long-range OD, where selectively recovering high-resolution local details can improve localization.

\paragraph{\textbf{Performance under Reduced Global Image Resolution.}}

\begin{table*}[t]
\centering
\caption{Global-image-scale-wise micro-averaged OD IoU comparison across object categories. Object Exist denotes IoU computed only on samples where the target object exists, Total includes negative samples, and visual-token usage is normalized by Qwen2.5 (Baseline) using full-size global images.}
\label{tab:OD_scale}
\resizebox{0.99\textwidth}{!}{%
\begin{tabular}{lcccccccc}
\toprule
\multirow{2}{*}{\textbf{Model}} 
& \multicolumn{2}{c}{\textbf{Excavators $\uparrow$}} 
& \multicolumn{2}{c}{\textbf{Rebars $\uparrow$}} 
& \multicolumn{2}{c}{\textbf{Workers w/ white hard hats $\uparrow$}}
& \multirow{2}{*}{\textbf{Avg. IoU $\uparrow$}}
& \multirow{2}{*}{\textbf{Visual Tokens $\downarrow$}} \\
\cmidrule(lr){2-3}
\cmidrule(lr){4-5}
\cmidrule(lr){6-7}
& \textbf{Object Exist} & \textbf{Total}
& \textbf{Object Exist} & \textbf{Total}
& \textbf{Object Exist} & \textbf{Total}
&  &  \\
\midrule
\midrule
\multicolumn{9}{c}{\textbf{Full-Size Global Image}} \\
Qwen2.5 (Baseline) (7B)~\cite{Qwen2_5}
& \underline{\textbf{87.0}} & \underline{\textbf{86.8}} 
& \underline{\textbf{47.3}} & \underline{\textbf{46.0}} 
& 72.1 & 69.3 
& \underline{\textbf{68.1}} 
& \underline{\textbf{100\%}}  \\

Qwen2.5$^{\star}$ (7B)~\cite{Qwen2_5}
& 84.7 & 79.0 
& 36.7 & 17.6 
& 60.9 & 59.3 
& 56.4  \textcolor{gray}{(-11.7$\downarrow$)}
& 193\% \textcolor{gray}{(+93\%$\uparrow$)} \\

AVA-VLM (7B) (Ours)
& 83.3 & 82.9 
& 45.0 & 42.0 
& \underline{\textbf{72.6}} & \underline{\textbf{69.4}} 
& 65.9  \textcolor{gray}{(-2.2$\downarrow$)}
& 101\% \textcolor{gray}{(+1\%$\uparrow$)} \\

\midrule
\multicolumn{9}{c}{\textbf{Global Image Downsampled to 1/2 Width and Height}} \\
Qwen2.5 (Baseline) (7B)~\cite{Qwen2_5}
& \underline{\textbf{82.9}} & \underline{\textbf{82.8}} 
& 35.8 & 35.4 
& 62.4 & 60.8 
& 60.0 
& \underline{\textbf{25\%}}  \\

Qwen2.5$^{\star}$ (7B)~\cite{Qwen2_5}
& 81.2 & 73.8 
& \underline{\textbf{39.2}} & 18.3 
& 53.8 & 52.7 
& 53.2 \textcolor{gray}{(-6.8$\downarrow$)}
& 48.3\% \textcolor{gray}{(+23.3\%$\uparrow$)}\\

AVA-VLM (7B) (Ours)
& 79.5 & 79.4 
& 38.3 & \underline{\textbf{37.4}} 
& \underline{\textbf{64.6}} & \underline{\textbf{62.6}} 
& \underline{\textbf{60.3}}  \textcolor{gray}{(+0.3$\uparrow$)}
& 25.8\% \textcolor{gray}{(+0.8\%$\uparrow$)} \\

\midrule
\multicolumn{9}{c}{\textbf{Global Image Downsampled to 1/4 Width and Height}} \\
Qwen2.5 (Baseline) (7B)~\cite{Qwen2_5}
& 72.5 & 72.0 
& 24.6 & 24.2 
& 43.3 & 42.3 
& 46.5 
& \underline{\textbf{6.3\%}} \\

Qwen2.5$^{\star}$ (7B)~\cite{Qwen2_5}
& 66.1 & 52.8 
& 25.9 & 10.9 
& 21.6 & 20.7 
& 33.0  \textcolor{gray}{(-13.5$\downarrow$)}
& 12.1\% \textcolor{gray}{(+5.8\%$\uparrow$)}\\

AVA-VLM (7B) (Ours)
& \underline{\textbf{73.5}} & \underline{\textbf{73.0}} 
& \underline{\textbf{29.5}} & \underline{\textbf{28.3}} 
& \underline{\textbf{52.4}} & \underline{\textbf{50.8}} 
& \underline{\textbf{51.3}}  \textcolor{gray}{(+4.8$\uparrow$)}
& 7.3\% \textcolor{gray}{(+1.0\%$\uparrow$)} \\

\bottomrule
\end{tabular}%
}
\end{table*}

Table~\ref{tab:OD_scale} evaluates OD performance under different global image resolutions. As the global image width and height are downsampled, Qwen2.5 (Baseline) shows a clear performance degradation, with the average IoU decreasing from 68.1\% with full-size global images to 46.5\% with 1/4-width-and-height global images. This confirms that OD is highly sensitive to global image resolution, because accurate bounding-box prediction requires fine-grained visual evidence for object boundaries. Qwen2.5$^{\star}$ also degrades under downsampling, with the average IoU decreasing from 56.4\% to 33.0\%. This indicates that example image-based CoT does not effectively compensate for the loss of visual details in the target image and can even interfere with localization.

In contrast, AVA-VLM shows stronger performance under reduced global image resolution. When the global image width and height are downsampled to 1/2, AVA-VLM achieves the highest average IoU of 60.3\%, slightly outperforming Qwen2.5 (Baseline) while using only 25.8\% of the full-size baseline visual-token budget. The advantage becomes more evident under the more aggressive 1/4-width-and-height setting, where AVA-VLM improves the average IoU from 46.5\% to 51.3\% over Qwen2.5 (Baseline), while using only 7.3\% visual tokens. These results suggest that adaptive local inspection becomes increasingly beneficial when the global image resolution is reduced, because the model can recover local details from the original image for cases where the downsampled global image is insufficient.

This behavior is also reflected in Figure~\ref{fig:OD_tool_call_ratio}(d). As the global image width and height are reduced, the tool-call ratio slightly decreases from 7.6\% with full-size global images to 5.9\% with 1/4-width-and-height global images. This indicates that severe downsampling can make some object evidence too unclear in the first-turn global observation, causing the model to more often answer directly without requesting local inspection. Therefore, AVA-VLM still experiences performance degradation as the global image becomes more aggressively downsampled. Nevertheless, when the model does trigger the crop tool, the recovered high-resolution local region helps preserve localization accuracy, leading to better reduced-resolution performance than the direct QA baseline.

Object-wise, AVA-VLM is particularly effective for smaller objects. For workers with white hard hats, AVA-VLM achieves the best Object Exist and Total IoU at all global image scales. This advantage is important because workers are smaller and more visually sensitive to downsampling than large objects such as excavators. Under the 1/4-width-and-height setting, AVA-VLM also achieves the best IoU across all three object categories, including excavators, rebars, and workers with white hard hats. These results demonstrate that AVA-VLM improves reduced-resolution OD performance by selectively recovering local details, with the benefit being especially pronounced for small-object localization.

\subsection{Image Captioning Performance Analysis}
\label{subsec:IC}

\begin{table*}[t]
\centering
\caption{Overall image captioning performance comparison on the IC task.}
\label{tab:constructionsite_captioning}
\resizebox{0.95\textwidth}{!}{%
\begin{tabular}{lcccc}
\toprule
\textbf{Model} & \textbf{SPICE $\uparrow$} & \textbf{METEOR $\uparrow$} & \textbf{BERTScore $\uparrow$} & \textbf{Average Words Per Caption} \\
\midrule

\multicolumn{5}{c}{\textbf{Zero-shot / Few-shot}} \\
GPT4V~\cite{GPT}            & 18.2 & 33.7 & 28.0 & 119 \textcolor{gray}{(+70)}\\
GPT4V 5-shot~\cite{GPT}     & 23.1 & 38.7 & 33.5 & 101 \textcolor{gray}{(+52)}\\
GPT4o 5-shot~\cite{GPT}     & 24.9 & \underline{\textbf{39.4}} & 37.7 & 80  \textcolor{gray}{(+31)}\\
MiniGPT4-v2~\cite{MiniGPT-v2}         & 15.5 & 28.3 & 31.1 & 60  \textcolor{gray}{(+11)}\\
LLaVA-v1.5 (7B)~\cite{LLaVA}         & 12.8 & 28.5 & 26.4 & 82  \textcolor{gray}{(+33)}\\
LLaVA-v1.5 (13B)~\cite{LLaVA}         & 14.6 & 29.8 & 27.8 & 84  \textcolor{gray}{(+35)}\\
LLaVA-v1.5 (13B) 5-shot~\cite{LLaVA}  & 15.1 & 30.8 & 27.2 & 101 \textcolor{gray}{(+52)}\\
Qwen2.5 (7B)~\cite{Qwen2_5}    & 17.4 & 34.5 & 31.0 & 90 \textcolor{gray}{(+41)}\\
\midrule

\multicolumn{5}{c}{\textbf{Construction-tailored VLMs (LoRA-tuning)}} \\
Qwen2.5 (Baseline) (7B)~\cite{Qwen2_5}     & 23.3 & 32.6 & 38.1 & 48.0 \textcolor{gray}{(-1.0)}\\
Qwen2.5$^{\star}$ (7B)~\cite{Qwen2_5}      & 20.9 & 30.2 & 37.2 & 39.5 \textcolor{gray}{(-9.5)}\\
AVA-VLM (7B) (Ours)                         & \underline{\textbf{25.6}} & 35.0 & \underline{\textbf{39.2}} & \underline{\textbf{49.6}} \textcolor{gray}{(+0.6)} \\

\midrule
Human annotation & --   & --   & --   & 49  \\
SCOCO SOTA       & 27.0 & 33.9 & --   & -- \\
\bottomrule
\end{tabular}%
}
\end{table*}

\label{subsec:IC}

In the proposed region-aware CoT dataset, IC annotations are designed as single-turn reasoning-and-description sequences because the task requires an overall scene-level description rather than query-specific safety-rule identification or object localization. As a result, IC annotations do not include tool calls, and AVA-VLM was observed to use the crop tool for 0\% of IC samples during inference.

Table~\ref{tab:constructionsite_captioning} compares image captioning performance on the IC task. Zero-shot and few-shot models tend to generate substantially longer captions than the construction-tailored VLMs. For example, GPT4V, GPT4V 5-shot, and Qwen2.5 produce 119, 101, and 90 words per caption on average, respectively, whereas the human annotations contain 49 words on average. These longer captions can increase lexical-overlap-based scores such as METEOR because METEOR rewards unigram matches through precision and recall. As a result, verbose captions have a higher chance of covering words or phrases that appear in the reference captions, which helps explain the strong METEOR scores of GPT4V 5-shot, GPT4o 5-shot, and Qwen2.5. 

In contrast, construction-tailored VLMs generate more compact descriptions that are closer to the human annotation length. Qwen2.5 (Baseline), Qwen2.5$^{\star}$, and AVA-VLM produce 48.0, 39.5, and 49.6 words per caption on average, respectively, closely matching the human annotation length of 49 words. This suggests that construction-specific LoRA tuning encourages the models to generate more focused and meaningful construction-site descriptions rather than overly verbose general-purpose captions.

Among the construction-tailored VLMs, AVA-VLM achieves the best overall captioning performance, with the highest SPICE, METEOR, and BERTScore scores of 25.6, 35.0, and 39.2, respectively. The higher SPICE score indicates that AVA-VLM better captures the semantic content of construction-site scenes. SPICE evaluates caption quality by parsing captions into scene-graph tuples that represent objects, attributes, and relations, and then comparing these semantic tuples with those extracted from reference captions. Therefore, the improvement in SPICE suggests that AVA-VLM more accurately describes key construction-site elements and their relationships, rather than simply generating longer captions with more words.

AVA-VLM also achieves the highest BERTScore among all compared models. Unlike lexical-overlap metrics, BERTScore measures semantic similarity using contextual token embeddings, making it more sensitive to whether the generated caption conveys a meaning similar to the reference even when the exact wording differs. The higher BERTScore of AVA-VLM therefore indicates stronger semantic alignment with human annotations. Together with its average caption length of 49.6 words, which is close to the human annotation length of 49 words, these results show that AVA-VLM generates captions that are both compact and semantically accurate.

%
%
%

\clearpage

\section{Conclusion}
In this study, we proposed AVA-VLM, an adaptive visual attention-based vision-language model for in-the-wild construction-site monitoring. Unlike conventional direct QA-based construction-tailored VLMs that generate answers from a high-resolution global image without explicitly deciding whether additional visual inspection is needed, AVA-VLM follows a coarse-to-fine reasoning strategy inspired by human visual inspection. The model first reasons over a downsampled global image to capture the overall construction-site context and then selectively requests a high-resolution local crop only when the query-relevant evidence is insufficient. This design enables AVA-VLM to preserve global scene understanding while adaptively recovering fine-grained visual details for small, distant, or visually ambiguous regions.

To train this behavior, we introduced a region-aware CoT dataset by extending direct QA-style construction-site annotations in ConstructionSite10K dataset into task-specific reasoning sequences for violation identification, object detection, and image captioning. For violation identification and object detection, the dataset explicitly supervises when local inspection is needed, where the model should crop, and how the cropped visual evidence should be incorporated into the final response. This annotation design allows the model to learn both direct-answer and tool-use behaviors within a unified construction-site VLM framework. Although our violation-identification experiments focus on PPE non-compliance due to the annotation coverage and the number of positive samples given by ConstructionSite10K~\cite{ConstructionSite10K}, the proposed framework is not restricted to this specific safety category. Given sufficient task-specific annotations, the same region-aware CoT annotation design and adaptive local-inspection mechanism can be transferred to other construction-safety violations and object categories without changing the overall training and inference pipeline.

Extensive experiments demonstrated the advantages of AVA-VLM across multiple construction-site monitoring tasks. For violation identification, AVA-VLM achieved the best overall classification performance while using substantially fewer visual tokens than both direct QA-based adaptation and example image-based CoT. It also improved reasoning and localization quality, especially for distant PPE-violation cases and reduced-resolution inputs, by selectively inspecting high-resolution local regions when global evidence was insufficient. For object detection, AVA-VLM provided a favorable performance--efficiency trade-off and showed clear advantages for long-distance and small-object detection. Under reduced global image resolution, AVA-VLM consistently outperformed direct QA-based and example image-based baselines, demonstrating the benefit of adaptive local inspection when global visual details are degraded. For image captioning, AVA-VLM generated compact, semantically accurate captions without invoking the crop tool, showing the superiority of AVA-VLM in global scene description.

Overall, the results show that adaptive visual attention is a promising direction for practical construction-site VLMs. By combining low-resolution global reasoning with selective high-resolution local inspection, AVA-VLM jointly addresses three key challenges in in-the-wild construction-site monitoring: limited operational range for small or distant objects, performance degradation under reduced-resolution inputs, and excessive visual-token consumption. 

\section{Limitation and future work}

Although this study addresses several critical challenges of VLMs for in-the-wild construction-site monitoring, including operational range, reduced-resolution reliability, and visual-token efficiency, several limitations remain. We view these limitations not as fundamental restrictions of the proposed adaptive visual attention framework, but as important directions for extending its applicability to broader construction-safety scenarios.

First, the current task coverage is limited by the available annotations in the source dataset. For violation identification, this study focuses on PPE non-compliance. ConstructionSite10K is one of the few construction-site VLM datasets that covers short-, mid-, and long-distance monitoring scenarios, making it suitable for studying operational range. However, although the original dataset also includes safety-harness, excavator-proximity, and edge-protection violations, these categories contain highly limited positive samples. Specifically, safety-harness and excavator-proximity violations contain fewer than 100 positive instances, while edge-protection violations contain fewer than 200 positive instances. Since AVA-VLM is trained through annotation-based supervised fine-tuning, these rare categories do not provide sufficient positive examples for training. Therefore, we excluded them from the current VI experiments and focused on PPE violations, which provide enough positive samples for meaningful training and evaluation. Similarly, for object detection, the current OD task is limited to the three object categories provided by ConstructionSite10K: excavators, rebars, and workers with white hard hats. As a result, the current evaluation does not fully cover the diversity of safety hazards, materials, and worker activities that appear in real construction sites.

It is important to note, however, that this limitation mainly arises from the annotation coverage of the source dataset rather than from the proposed framework itself. AVA-VLM is designed to be dataset-agnostic in the sense that it does not rely on task-specific model components or specific safety categories. Given sufficient task-specific annotations, the same region-aware CoT annotation generation process can be applied to other violation types or object categories. Therefore, once adequate data are available, the proposed adaptive visual attention framework can be easily transferred to broader construction-safety monitoring tasks without changing the overall training and inference design.

More fundamentally, to address this coverage limitation, future work will explore reinforcement-learning-based policy optimization with automatically generated supervision. Instead of relying only on human-labeled annotations for each safety category, a VLM can generate multiple rollouts for unlabeled construction-site images, including different crop decisions, candidate crop regions, reasoning sequences, and final answers. Strong multimodal models, such as GPT or Gemini, can then be used as judges to evaluate whether a crop is needed, whether the selected crop region is appropriate, and whether the final answer is supported by the visual evidence. These judgments can provide reward signals for training the model's adaptive visual attention policy. Such an approach could expand the coverage of AVA-VLM to more safety categories and object types with substantially less human labeling effort. Moreover, by exposing the model to diverse unlabeled images and rollout, reinforcement learning may improve robustness to edge cases and out-of-distribution construction-site conditions that are difficult to capture with fixed supervised annotations.

Second, the current framework is image-based, which is also a common limitation of most existing construction-site VLM studies~\cite{ConstructionVLM1, ConstructionVLM2, ConstructionVCSQ, ConstructionContext-aware-vlm, ConstructionCS-VLM, ConstructionDoubleThinking, ConstructionEnhancingVLM_VGR_RL}. Existing construction-tailored VLMs have primarily focused on image-level understanding tasks without explicitly modeling temporal dynamics across frames. Our study also follows this image-based setting. However, in real construction-site monitoring, safety-related events often evolve over time. For example, unsafe movement patterns or temporary PPE removal may be difficult to determine from a single image but become clearer when multiple frames are observed. The proposed adaptive visual attention mechanism currently decides whether and where to inspect based only on the current image and query. Extending AVA-VLM to video-based monitoring would allow the current spatial zoom-in tool to be combined with temporal tracking. For example, the model could track workers, equipment, or safety-critical regions across frames and selectively trigger high-resolution local inspection when temporal evidence indicates increasing risk or visual ambiguity, such as when a worker approaches an excavator or when PPE visibility becomes uncertain over consecutive frames. Future work will therefore investigate temporal adaptive visual attention, where the model can decide not only where to inspect spatially, but also when to inspect temporally, providing a clearer path toward efficient real-world CCTV-based construction-site monitoring.

Third, although AVA-VLM substantially reduces visual-token usage from 100\% to 30.6\% in PPE-violation identification, CoT-based reasoning still introduces additional language-token overhead by generating intermediate reasoning traces. Since VLM inference time depends not only on visual-token processing but also on language-token generation, long reasoning traces can reduce the practical efficiency gained from visual-token reduction, especially for real-time edge deployment. In this study, the reasoning traces are designed to make the model's decision process explicit and to supervise when local inspection is necessary. However, for real-time deployment, it may be beneficial to generate more concise reasoning or to internally perform reasoning while outputting only a compact final response. Future work will investigate methods for compressing generated reasoning, such as shorter tool-use rationales or distillation from verbose CoT traces into more concise decision policies, to further optimize end-to-end latency for real-time edge deployment.

These limitations suggest several promising directions for future research. Expanding task coverage with reinforcement-learning-based training, incorporating temporal context for video-based monitoring, and reducing the language-generation overhead of CoT reasoning can further improve the practicality of AVA-VLM. Addressing these challenges will help move construction-site VLMs closer to reliable, efficient, and scalable deployment in real-world monitoring environments.

\section*{CRediT authorship contribution statement}
\textbf{Younggun Kim:} Writing – review \& editing, Writing – original draft,
Visualization, Validation, Methodology, Formal
analysis, Data curation, Conceptualization. \textbf{Taeheon Kim:} Writing –
review \& editing, Validation. \textbf{Youngseo Kim:} Writing –
review \& editing, Supervision. 
\textbf{Seunghee Park:} Writing –
review \& editing, Supervision, Resources, Project administration, Funding
acquisition.

\section*{Declaration of competing interest} 
The authors declare the following financial interests/personal relationships which may be considered as potential competing interests: Seunghee Park reports financial support was provided by Sungkyunkwan University.

\section*{Acknowledgments}
This work was supported by the National Research Foun dation of Korea(NRF) grant funded by the Korea govern ment(MSIT) (RS-2025-02223612, RS-2024-00336270).

During the preparation of this work the authors used GPT-5.5 in order to grammar check and language editing. After using this tool, the authors reviewed and edited the content as needed and take full responsibility for the content of the article.

\section*{Data availability}
Data will be made available on request.


\bibliographystyle{cas-model2-names}

\bibliography{cas-refs}

@misc{virtualassistants1,
      title={Intelligent Virtual Assistants with LLM-based Process Automation}, 
      author={Yanchu Guan and Dong Wang and Zhixuan Chu and Shiyu Wang and Feiyue Ni and Ruihua Song and Longfei Li and Jinjie Gu and Chenyi Zhuang},
      year={2023},
      journal = {arXiv preprint arXiv:2312.06677},
}

@misc{virtualassistants2,
      title={SELMA: A Speech-Enabled Language Model for Virtual Assistant Interactions}, 
      author={Dominik Wagner and Alexander Churchill and Siddharth Sigtia and Erik Marchi},
      year={2025},
      journal = {arXiv preprint arXiv:2501.19377},
}

@article{privacy1,
  title={Privacy-aware visual language models},
  author={Samson, Laurens and Barazani, Nimrod and Ghebreab, Sennay and Asano, Yuki M},
  journal={arXiv preprint arXiv:2405.17423},
  year={2024}
}

@misc{privacy2,
      title={Safe-LLaVA: A Privacy-Preserving Vision-Language Dataset and Benchmark for Biometric Safety}, 
      author={Younggun Kim and Sirnam Swetha and Fazil Kagdi and Mubarak Shah},
      year={2025},
      eprint={2509.00192},
      archivePrefix={arXiv},
      primaryClass={cs.CV},
      url={https://arxiv.org/abs/2509.00192}, 
}

@misc{transportation1,
      title={VRU-Accident: A Vision-Language Benchmark for Video Question Answering and Dense Captioning for Accident Scene Understanding}, 
      author={Kim, Younggun and Abdelrahman, Ahmed S. and Abdel-Aty, Mohamed},
      year={2025},
      journal = {arXiv preprint arXiv:2507.09815},
}

@ARTICLE{transportation2,
  author={Tran, Dai Quoc and Abdel-Aty, Mohamed and Kim, Younggun and Abdelrahman, Ahmed S. and Islam, Zubayer},
  journal={IEEE Transactions on Intelligent Transportation Systems}, 
  title={Region-Level Vision-Language Model for Detecting Distraction Behavior and Mobility Attributes of Vulnerable Road Users}, 
  year={2026},
  volume={27},
  number={5},
  pages={5784-5803},
  doi={10.1109/TITS.2026.3657271}}

@InProceedings{transportation3,
    author    = {AlShami, Ali K. and Rabinowitz, Ryan and Shoman, Maged and Fang, Jianwu and Picek, Lukas and Lo, Shao-Yuan and Cruz, Steve and Lam, Khang Nhut and Kamod, Nachiket and Li, Lei-Lei and Kalita, Jugal and Boult, Terrance E.},
    title     = {2COOOL: 2nd Workshop on the Challenge Of Out-Of-Label Hazards in Autonomous Driving},
    booktitle = {Proceedings of the IEEE/CVF International Conference on Computer Vision (ICCV) Workshops},
    month     = {October},
    year      = {2025},
    pages     = {764-771}
}

@article{ConstructionTran1,
title = {Visual Question Answering-based Referring Expression Segmentation for construction safety analysis},
journal = {Automation in Construction},
volume = {174},
pages = {106127},
year = {2025},
issn = {0926-5805},
doi = {https://doi.org/10.1016/j.autcon.2025.106127},
url = {https://www.sciencedirect.com/science/article/pii/S0926580525001670},
author = {Dai Quoc Tran and Armstrong Aboah and Yuntae Jeon and Minh-Truyen Do and Mohamed Abdel-Aty and Minsoo Park and Seunghee Park}
}

@article{ConstructionDoubleThinking,
  title={A Double Thinking Enabled Visual Language Model for Open-Set Construction Site Safety Inspections},
  author={Tang, Yutong and Yan, Hui and Gao, Zeyu and Zhang, Zhen and Luo, Xiaochun},
  journal={Available at SSRN 5179874}
}

@article{ConstructionEnhancingVLM_VGR_RL,
  title={Enhancing Vision-Language Model for Construction Safety Inspection via Visually Grounded Reasoning and Reinforcement Learning},
  author={Chan, Jeff Chak Fu and WONG, Peter Kok-Yiu and Guo, Xiaowen and Liang, Zhenyu and Wu, Yifan and Cheng, Jack CP},
  journal={Available at SSRN 6294267}
}

@article{ConstructionContext-aware-vlm,
  title={Context-aware vision-language model agent enriched with domain-specific ontology for construction site safety monitoring},
  author={Chan, Chak-Fu and Wong, Peter Kok-Yiu and Guo, Xiaowen and Cheng, Jack CP and Chan, Jolly Pui-Ching and Leung, Pak-Him and Tao, Xingyu},
  journal={Automation in Construction},
  volume={177},
  pages={106305},
  year={2025},
  publisher={Elsevier}
}

@inproceedings{Resolution1,
  title={Res-Bench: Benchmarking the Robustness of Multimodal Large Language Models to Dynamic Resolution Input},
  author={Li, Chenxu and Wang, Zhicai and Sheng, Yuan and Zhu, Xingyu and Hao, Yanbin and Wang, Xiang},
  booktitle={Proceedings of the AAAI Conference on Artificial Intelligence},
  volume={40},
  number={37},
  pages={31545--31553},
  year={2026}
}

@misc{Resolution2,
      title={VLM-RobustBench: A Comprehensive Benchmark for Robustness of Vision-Language Models}, 
      author={Rohit Saxena and Alessandro Suglia and Pasquale Minervini},
      year={2026},
      eprint={2603.06148},
      archivePrefix={arXiv},
      primaryClass={cs.CV},
      url={https://arxiv.org/abs/2603.06148}, 
}

@article{Edge1,
  title={Vision-language models for edge networks: A comprehensive survey},
  author={Sharshar, Ahmed and Khan, Latif U and Ullah, Waseem and Guizani, Mohsen},
  journal={IEEE Internet of Things Journal},
  year={2025},
  publisher={IEEE}
}

@article{Edge2,
  title={Edge video analytics: A survey on applications, systems and enabling techniques},
  author={Xu, Renjie and Razavi, Saiedeh and Zheng, Rong},
  journal={IEEE Communications Surveys \& Tutorials},
  volume={25},
  number={4},
  pages={2951--2982},
  year={2023},
  publisher={IEEE}
}

@article{CoT_for_Generalization1,
  title={Multimodal chain-of-thought reasoning in language models},
  author={Zhang, Zhuosheng and Zhang, Aston and Li, Mu and Zhao, Hai and Karypis, George and Smola, Alex},
  journal={arXiv preprint arXiv:2302.00923},
  year={2023}
}

@article{CoT_for_Generalization2,
  title={Chain of thought prompt tuning in vision language models},
  author={Ge, Jiaxin and Luo, Hongyin and Qian, Siyuan and Gan, Yulu and Fu, Jie and Zhang, Shanghang},
  journal={arXiv preprint arXiv:2304.07919},
  year={2023}
}

@article{ConstructionCS-VLM,
title = {Construction site fall hazard identification and automated captioning using adapted vision-language models},
journal = {Automation in Construction},
volume = {183},
pages = {106790},
year = {2026},
issn = {0926-5805},
doi = {https://doi.org/10.1016/j.autcon.2026.106790},
url = {https://www.sciencedirect.com/science/article/pii/S0926580526000312},
author = {Yongshuang Li and Feng Xu and Zhipeng Zhang and Xinyu Mei and He Huang}
}

@article{ConstructionVLMplusYOLO,
title = {Real-time safety detection on construction sites using a vision-language and NLP-based model},
journal = {Advanced Engineering Informatics},
volume = {69},
pages = {103889},
year = {2026},
issn = {1474-0346},
doi = {https://doi.org/10.1016/j.aei.2025.103889},
url = {https://www.sciencedirect.com/science/article/pii/S1474034625007827},
author = {S. Sivanraj and D.N.L.S Uduwage and M. Tripathi}
}

@article{ConstructionVCSQ,
title = {Augmented reality, deep learning and vision-language query system for construction worker safety},
journal = {Automation in Construction},
volume = {157},
pages = {105158},
year = {2024},
issn = {0926-5805},
doi = {https://doi.org/10.1016/j.autcon.2023.105158},
url = {https://www.sciencedirect.com/science/article/pii/S0926580523004181},
author = {Haosen Chen and Lei Hou and Shaoze Wu and Guomin Zhang and Yang Zou and Sungkon Moon and Muhammed Bhuiyan}
}

@article{ConstructionVLM1,
title = {Tailored vision-language framework for automated hazard identification and report generation in construction sites},
journal = {Advanced Engineering Informatics},
volume = {66},
pages = {103478},
year = {2025},
issn = {1474-0346},
doi = {https://doi.org/10.1016/j.aei.2025.103478},
url = {https://www.sciencedirect.com/science/article/pii/S1474034625003714},
author = {Qihua Chen and Xianfei Yin}
}

@article{ConstructionVLM2,
title = {Vision transformer-based visual language understanding of the construction process},
journal = {Alexandria Engineering Journal},
volume = {99},
pages = {242-256},
year = {2024},
issn = {1110-0168},
doi = {https://doi.org/10.1016/j.aej.2024.05.015},
url = {https://www.sciencedirect.com/science/article/pii/S1110016824004873},
author = {Bin Yang and Binghan Zhang and Yilong Han and Boda Liu and Jiniming Hu and Yiming Jin}
}

@misc{LoRA,
      title={LoRA: Low-Rank Adaptation of Large Language Models}, 
      author={Edward J. Hu and Yelong Shen and Phillip Wallis and Zeyuan Allen-Zhu and Yuanzhi Li and Shean Wang and Lu Wang and Weizhu Chen},
      year={2021},
      eprint={2106.09685},
      archivePrefix={arXiv},
      primaryClass={cs.CL},
      url={https://arxiv.org/abs/2106.09685}, 
}

@article{ConstructionSite10K,
   title={Are large pre-trained vision language models effective construction safety inspectors},
   volume={7},
   ISSN={2632-6736},
   url={http://dx.doi.org/10.1017/dce.2026.10044},
   DOI={10.1017/dce.2026.10044},
   journal={Data-Centric Engineering},
   publisher={Cambridge University Press (CUP)},
   author={Chen, Xuezheng and Zou, Zhengbo},
   year={2026} }

@misc{AdaptVision,
      title={AdaptVision: Efficient Vision-Language Models via Adaptive Visual Acquisition}, 
      author={Zichuan Lin and Yicheng Liu and Yang Yang and Lvfang Tao and Deheng Ye},
      year={2026},
      eprint={2512.03794},
      archivePrefix={arXiv},
      primaryClass={cs.CV},
      url={https://arxiv.org/abs/2512.03794}, 
}

@article{VisionThink,
  title={Visionthink: Smart and efficient vision language model via reinforcement learning},
  author={Yang, Senqiao and Li, Junyi and Lai, Xin and Wu, Jinming and Li, Wei and MA, Zejun and Yu, Bei and Zhao, Hengshuang and Jia, Jiaya},
  journal={Advances in Neural Information Processing Systems},
  volume={38},
  pages={95187--95227},
  year={2026}
}

@inproceedings{LLaVA,
 author = {Liu, Haotian and Li, Chunyuan and Wu, Qingyang and Lee, Yong Jae},
 booktitle = {Advances in Neural Information Processing Systems},
 editor = {A. Oh and T. Naumann and A. Globerson and K. Saenko and M. Hardt and S. Levine},
 pages = {34892--34916},
 publisher = {Curran Associates, Inc.},
 title = {Visual Instruction Tuning},
 url = {https://proceedings.neurips.cc/paper_files/paper/2023/file/6dcf277ea32ce3288914faf369fe6de0-Paper-Conference.pdf},
 volume = {36},
 year = {2023}
}

@misc{LLaVA-NeXT,
    title={LLaVA-NeXT: Improved reasoning, OCR, and world knowledge},
    url={https://llava-vl.github.io/blog/2024-01-30-llava-next/},
    author={Liu, Haotian and Li, Chunyuan and Li, Yuheng and Li, Bo and Zhang, Yuanhan and Shen, Sheng and Lee, Yong Jae},
    year={2024}
}

@inproceedings{LLaVA-CoT,
  title={Llava-cot: Let vision language models reason step-by-step},
  author={Xu, Guowei and Jin, Peng and Wu, Ziang and Li, Hao and Song, Yibing and Sun, Lichao and Yuan, Li},
  booktitle={Proceedings of the IEEE/CVF International Conference on Computer Vision},
  year={2025}
}

@misc{QwenVL,
      title={Qwen-VL: A Versatile Vision-Language Model for Understanding, Localization, Text Reading, and Beyond}, 
      author={Jinze Bai and Shuai Bai and Shusheng Yang and Shijie Wang and Sinan Tan and Peng Wang and Junyang Lin and Chang Zhou and Jingren Zhou},
      year={2023},
      journal = {arXiv preprint arXiv:2308.12966},
}

@misc{Qwen2VL,
      title={Qwen2-VL: Enhancing Vision-Language Model's Perception of the World at Any Resolution}, 
      author={Peng Wang and Shuai Bai and Sinan Tan and Shijie Wang and Zhihao Fan and Jinze Bai and Keqin Chen and Xuejing Liu and Jialin Wang and Wenbin Ge and Yang Fan and Kai Dang and Mengfei Du and Xuancheng Ren and Rui Men and Dayiheng Liu and Chang Zhou and Jingren Zhou and Junyang Lin},
      year={2024},
     journal={arXiv preprint arXiv:2409.12191},
}

@article{Qwen2_5,
    title   = {Qwen2.5 Technical Report}, 
    author  = {An Yang and Baosong Yang and Beichen Zhang and Binyuan Hui and Bo Zheng and Bowen Yu and Chengyuan Li and Dayiheng Liu and Fei Huang and Haoran Wei and Huan Lin and Jian Yang and Jianhong Tu and Jianwei Zhang and Jianxin Yang and Jiaxi Yang and Jingren Zhou and Junyang Lin and Kai Dang and Keming Lu and Keqin Bao and Kexin Yang and Le Yu and Mei Li and Mingfeng Xue and Pei Zhang and Qin Zhu and Rui Men and Runji Lin and Tianhao Li and Tingyu Xia and Xingzhang Ren and Xuancheng Ren and Yang Fan and Yang Su and Yichang Zhang and Yu Wan and Yuqiong Liu and Zeyu Cui and Zhenru Zhang and Zihan Qiu},
    journal = {arXiv preprint arXiv:2412.15115},
    year    = {2024}
}

@misc{InternVL2_5,
      title={Expanding Performance Boundaries of Open-Source Multimodal Models with Model, Data, and Test-Time Scaling}, 
      author={Zhe Chen and Weiyun Wang and Yue Cao and Yangzhou Liu and Zhangwei Gao and Erfei Cui and Jinguo Zhu and Shenglong Ye and Hao Tian and Zhaoyang Liu and Lixin Gu and Xuehui Wang and Qingyun Li and Yimin Ren and Zixuan Chen and Jiapeng Luo and Jiahao Wang and Tan Jiang and Bo Wang and Conghui He and Botian Shi and Xingcheng Zhang and Han Lv and Yi Wang and Wenqi Shao and Pei Chu and Zhongying Tu and Tong He and Zhiyong Wu and Huipeng Deng and Jiaye Ge and Kai Chen and Kaipeng Zhang and Limin Wang and Min Dou and Lewei Lu and Xizhou Zhu and Tong Lu and Dahua Lin and Yu Qiao and Jifeng Dai and Wenhai Wang},
      year={2025},
      journal={arXiv preprint arXiv:2412.05271},
}

@misc{InternVL3,
      title={InternVL3: Exploring Advanced Training and Test-Time Recipes for Open-Source Multimodal Models}, 
      author={Jinguo Zhu and Weiyun Wang and Zhe Chen and Zhaoyang Liu and Shenglong Ye and Lixin Gu and Hao Tian and Yuchen Duan and Weijie Su and Jie Shao and Zhangwei Gao and Erfei Cui and Xuehui Wang and Yue Cao and Yangzhou Liu and Xingguang Wei and Hongjie Zhang and Haomin Wang and Weiye Xu and Hao Li and Jiahao Wang and Nianchen Deng and Songze Li and Yinan He and Tan Jiang and Jiapeng Luo and Yi Wang and Conghui He and Botian Shi and Xingcheng Zhang and Wenqi Shao and Junjun He and Yingtong Xiong and Wenwen Qu and Peng Sun and Penglong Jiao and Han Lv and Lijun Wu and Kaipeng Zhang and Huipeng Deng and Jiaye Ge and Kai Chen and Limin Wang and Min Dou and Lewei Lu and Xizhou Zhu and Tong Lu and Dahua Lin and Yu Qiao and Jifeng Dai and Wenhai Wang},
      year={2025},
      journal={arXiv preprint arXiv:2504.10479},
}

@misc{InternVL3_5,
      title={InternVL3.5: Advancing Open-Source Multimodal Models in Versatility, Reasoning, and Efficiency}, 
      author={Weiyun Wang and Zhangwei Gao and Lixin Gu and Hengjun Pu and Long Cui and Xingguang Wei and Zhaoyang Liu and Linglin Jing and Shenglong Ye and Jie Shao and Zhaokai Wang and Zhe Chen and Hongjie Zhang and Ganlin Yang and Haomin Wang and Qi Wei and Jinhui Yin and Wenhao Li and Erfei Cui and Guanzhou Chen and Zichen Ding and Changyao Tian and Zhenyu Wu and Jingjing Xie and Zehao Li and Bowen Yang and Yuchen Duan and Xuehui Wang and Zhi Hou and Haoran Hao and Tianyi Zhang and Songze Li and Xiangyu Zhao and Haodong Duan and Nianchen Deng and Bin Fu and Yinan He and Yi Wang and Conghui He and Botian Shi and Junjun He and Yingtong Xiong and Han Lv and Lijun Wu and Wenqi Shao and Kaipeng Zhang and Huipeng Deng and Biqing Qi and Jiaye Ge and Qipeng Guo and Wenwei Zhang and Songyang Zhang and Maosong Cao and Junyao Lin and Kexian Tang and Jianfei Gao and Haian Huang and Yuzhe Gu and Chengqi Lyu and Huanze Tang and Rui Wang and Haijun Lv and Wanli Ouyang and Limin Wang and Min Dou and Xizhou Zhu and Tong Lu and Dahua Lin and Jifeng Dai and Weijie Su and Bowen Zhou and Kai Chen and Yu Qiao and Wenhai Wang and Gen Luo},
      year={2025},
      eprint={2508.18265},
      archivePrefix={arXiv},
      primaryClass={cs.CV},
      url={https://arxiv.org/abs/2508.18265}, 
}

@misc{MiniGPT-v2,
      title={MiniGPT-v2: large language model as a unified interface for vision-language multi-task learning}, 
      author={Jun Chen and Deyao Zhu and Xiaoqian Shen and Xiang Li and Zechun Liu and Pengchuan Zhang and Raghuraman Krishnamoorthi and Vikas Chandra and Yunyang Xiong and Mohamed Elhoseiny},
      year={2023},
      eprint={2310.09478},
      archivePrefix={arXiv},
      primaryClass={cs.CV},
      url={https://arxiv.org/abs/2310.09478}, 
}

@misc{GPT,
      title={GPT-4 Technical Report}, 
      author={OpenAI},
      year={2024},
      eprint={2303.08774},
      archivePrefix={arXiv},
      primaryClass={cs.CL},
      url={https://arxiv.org/abs/2303.08774}, 
}

@article{GroundingDino,
  title={Grounding dino: Marrying dino with grounded pre-training for open-set object detection},
  author={Liu, Shilong and Zeng, Zhaoyang and Ren, Tianhe and Li, Feng and Zhang, Hao and Yang, Jie and Li, Chunyuan and Yang, Jianwei and Su, Hang and Zhu, Jun and others},
  journal={arXiv preprint arXiv:2303.05499},
  year={2023}
}

@inproceedings{visualtoken1,
  title={Visionzip: Longer is better but not necessary in vision language models},
  author={Yang, Senqiao and Chen, Yukang and Tian, Zhuotao and Wang, Chengyao and Li, Jingyao and Yu, Bei and Jia, Jiaya},
  booktitle={Proceedings of the IEEE/CVF Conference on Computer Vision and Pattern Recognition},
  pages={19792--19802},
  year={2025}
}

@article{visualtoken2,
  title={Efficient vision-language models by summarizing visual tokens into compact registers},
  author={Wen, Yuxin and Cao, Qingqing and Fu, Qichen and Mehta, Sachin and Najibi, Mahyar},
  journal={arXiv preprint arXiv:2410.14072},
  year={2024}
}

@inproceedings{visualtoken3,
  title={Upop: Unified and progressive pruning for compressing vision-language transformers},
  author={Shi, Dachuan and Tao, Chaofan and Jin, Ying and Yang, Zhendong and Yuan, Chun and Wang, Jiaqi},
  booktitle={International Conference on Machine Learning},
  pages={31292--31311},
  year={2023},
  organization={PMLR}
}

@inproceedings{visualtoken4,
  title={Expedited training of visual conditioned language generation via redundancy reduction},
  author={Jian, Yiren and Liu, Tingkai and Tao, Yunzhe and Zhang, Chunhui and Vosoughi, Soroush and Yang, Hongxia},
  booktitle={Proceedings of the 62nd Annual Meeting of the Association for Computational Linguistics (Volume 1: Long Papers)},
  pages={300--314},
  year={2024}
}

@article{visualtoken5,
  title={Sparsevlm: Visual token sparsification for efficient vision-language model inference},
  author={Zhang, Yuan and Fan, Chun-Kai and Ma, Junpeng and Zheng, Wenzhao and Huang, Tao and Cheng, Kuan and Gudovskiy, Denis and Okuno, Tomoyuki and Nakata, Yohei and Keutzer, Kurt and others},
  journal={arXiv preprint arXiv:2410.04417},
  year={2024}
}

@misc{visualtoken6,
      title={An Image is Worth 1/2 Tokens After Layer 2: Plug-and-Play Inference Acceleration for Large Vision-Language Models}, 
      author={Liang Chen and Haozhe Zhao and Tianyu Liu and Shuai Bai and Junyang Lin and Chang Zhou and Baobao Chang},
      year={2024},
      eprint={2403.06764},
      archivePrefix={arXiv},
      primaryClass={cs.CV},
      url={https://arxiv.org/abs/2403.06764}, 
}

@misc{visualtoken7,
      title={PyramidDrop: Accelerating Your Large Vision-Language Models via Pyramid Visual Redundancy Reduction}, 
      author={Long Xing and Qidong Huang and Xiaoyi Dong and Jiajie Lu and Pan Zhang and Yuhang Zang and Yuhang Cao and Conghui He and Jiaqi Wang and Feng Wu and Dahua Lin},
      year={2025},
      eprint={2410.17247},
      archivePrefix={arXiv},
      primaryClass={cs.CV},
      url={https://arxiv.org/abs/2410.17247}, 
}

@misc{llama,
      title={The Llama 3 Herd of Models}, 
      author={Meta},
      year={2024},
      eprint={2407.21783},
      archivePrefix={arXiv},
      primaryClass={cs.AI},
      url={https://arxiv.org/abs/2407.21783}, 
}

@inproceedings{MMICL,
  title={Mmicl: Empowering vision-language model with multi-modal in-context learning},
  author={Zhao, Haozhe and Cai, Zefan and Si, Shuzheng and Ma, Xiaojian and An, Kaikai and Chen, Liang and Liu, Zixuan and Wang, Sheng and Han, Wenjuan and Chang, Baobao},
  booktitle={International Conference on Learning Representations},
  volume={2024},
  pages={14942--14980},
  year={2024}
}

@inproceedings{MultimodalCoT,
  title={Can Multimodal Large Language Models Truly Perform Multimodal In-Context Learning?},
  author={Chen, Shuo and Han, Zhen and He, Bailan and Liu, Jianzhe and Buckley, Mark and Qin, Yao and Torr, Philip and Tresp, Volker and Gu, Jindong},
  booktitle={2025 IEEE/CVF Winter Conference on Applications of Computer Vision (WACV)},
  pages={6000--6010},
  year={2025},
  organization={IEEE}
}

@inproceedings{SPICE,
  title={Spice: Semantic propositional image caption evaluation},
  author={Anderson, Peter and Fernando, Basura and Johnson, Mark and Gould, Stephen},
  booktitle={European conference on computer vision},
  pages={382--398},
  year={2016},
  organization={Springer}
}

@inproceedings{METEOR,
  title={METEOR: An automatic metric for MT evaluation with improved correlation with human judgments},
  author={Banerjee, Satanjeev and Lavie, Alon},
  booktitle={Proceedings of the acl workshop on intrinsic and extrinsic evaluation measures for machine translation and/or summarization},
  pages={65--72},
  year={2005}
}

@article{BertScore,
  title={Bertscore: Evaluating text generation with bert},
  author={Zhang, Tianyi and Kishore, Varsha and Wu, Felix and Weinberger, Kilian Q and Artzi, Yoav},
  journal={arXiv preprint arXiv:1904.09675},
  year={2019}
}

@article{VisualCoT,
  title={Visual cot: Advancing multi-modal language models with a comprehensive dataset and benchmark for chain-of-thought reasoning},
  author={Shao, Hao and Qian, Shengju and Xiao, Han and Song, Guanglu and Zong, Zhuofan and Wang, Letian and Liu, Yu and Li, Hongsheng},
  journal={Advances in Neural Information Processing Systems},
  volume={37},
  pages={8612--8642},
  year={2024}
}

@inproceedings{CropVLM,
  title={Cropvlm: Learning to zoom for fine-grained vision-language perception},
  author={Carvalho, Miguel and Dias, H{\'e}lder and Martins, Bruno},
  booktitle={Proceedings of the IEEE/CVF Conference on Computer Vision and Pattern Recognition},
  pages={11170--11179},
  year={2026}
}




\end{document}